\documentclass[lettersize,journal]{IEEEtran}

\usepackage{amsmath,amsfonts,amssymb}
\usepackage{algorithmic}
\usepackage{algorithm}
\usepackage{array}
\usepackage[caption=false,font=normalsize,labelfont=sf,textfont=sf]{subfig}
\usepackage{textcomp}
\usepackage{stfloats}
\usepackage{url}
\usepackage{graphicx}
\usepackage{cite}

\usepackage{booktabs}
\usepackage{colortbl}
\usepackage{makecell}
\usepackage{multirow}
\usepackage{xcolor}
\usepackage[hidelinks]{hyperref}
\usepackage[capitalise,noabbrev]{cleveref}

\newcommand{\eg}{\emph{e.g.}}
\newcommand{\ie}{\emph{i.e.}}
\newcommand{\etal}{\emph{et al.}}

\definecolor{headergray}{rgb}{0.92, 0.92, 0.92}
\definecolor{highlightblue}{rgb}{0.89, 0.89, 0.99}
\definecolor{highlightred}{rgb}{0.99, 0.89, 0.89}
\definecolor{positive}{rgb}{0.8, 0, 0}
\definecolor{negative}{rgb}{0, 0.5, 0}
\newcommand{\post}[1]{\textcolor{positive}{(#1)}}
\newcommand{\negt}[1]{\textcolor{negative}{(#1)}}

\crefname{figure}{Fig.}{Figs.}
\Crefname{figure}{Fig.}{Figs.}
\crefname{table}{Table}{Tables}
\Crefname{table}{Table}{Tables}
\crefname{section}{Section}{Sections}
\Crefname{section}{Section}{Sections}
\crefname{equation}{Eq.}{Eqs.}
\Crefname{equation}{Eq.}{Eqs.}

\begin{document}

\title{On the Adversarial Robustness of Multimodal LLM Judges}

\author{Zihan Wang,
        Guansong Pang*,~\IEEEmembership{Member,~IEEE},
        Zelin Liu,
        Wenjun Miao,
        Jin Zheng*,
        and Xiao Bai,~\IEEEmembership{Senior Member,~IEEE}

\thanks{*Corresponding author: Guansong Pang and Jin Zheng.}%
\thanks{Zihan Wang, Zelin Liu, and Wenjun Miao are with the School of Computer Science and Engineering, Beihang University, Beijing, China. (E-mail:\{wangzihan1118, liuzelin, miaowenjun\}@buaa.edu.cn).}%
\thanks{Jin Zheng is with the School of Computer Science and Engineering, Beihang University, Beijing, China and State Key Laboratory of Virtual Reality Technology and System, Beihang University, Beijing, China. (E-mail: jinzheng@buaa.edu.cn).}%
\thanks{Xiao Bai is with the School of Computer Science and Engineering, Beihang University, Beijing, China and State Key Laboratory of Software Development Environment, Jiangxi Research Institute, Beihang University, Beijing, China. (E-mail: baixiao@buaa.edu.cn).}%
\thanks{Guansong Pang is with the School of Computing and Information Systems,
Singapore Management University,  Singapore. (E-mail: gspang@smu.edu.sg)}

}


\maketitle

\begin{abstract}

Multimodal Large Language Models (MLLMs) are increasingly used as automated judges, \eg, for image quality and safety assessment.
However, their adversarial robustness remains largely unexplored, threatening the fairness and reliability of automated judging. To bridge this gap, we introduce \textbf{RobustMLLMJudge}, the first general framework for evaluating the adversarial robustness of general-purpose MLLMs when functioning as judges. It covers diverse attacks against popular judge approaches across quality and safety evaluation scenarios.
Using RobustMLLMJudge, we reveal that \textbf{i)} different MLLM judges are highly vulnerable to score-inflating adversarial attacks; and \textbf{ii)} although effective, these attack methods face a critical challenge due to unique constraints in the evaluation protocols of MLLM judges. We further propose \textbf{MGSIA}, namely \textbf{Manifold-Guided Semantic Induction Attack},
a novel method that bypasses these constraints to enable more effective and transferable attacks on MLLM judges. 
The core idea of MGSIA is to combine affirmative semantic induction with high-score manifold alignment: it maximizes the probability that judges yield affirmative responses (\eg, ``\texttt{Yes}'') to binary semantic queries, while regularizing adversarial representations toward high-score centers estimated from proxy protocols. 
Together, these objectives yield transferable score-inflating perturbations.
Extensive experiments demonstrate the superiority and generalizability of MGSIA in deceiving advanced MLLM judges under different evaluation scenarios, highlighting the need for robust MLLM judges. Code and data will be made available at \url{https://github.com/mala-lab/RobustMLLMJudge}.

\end{abstract}

\begin{IEEEkeywords}
Adversarial robustness, multimodal large language models, MLLM-as-a-judge, automated evaluation.
\end{IEEEkeywords}

\begin{figure}[!t]
  \centering
   \includegraphics[width=1.0\linewidth]{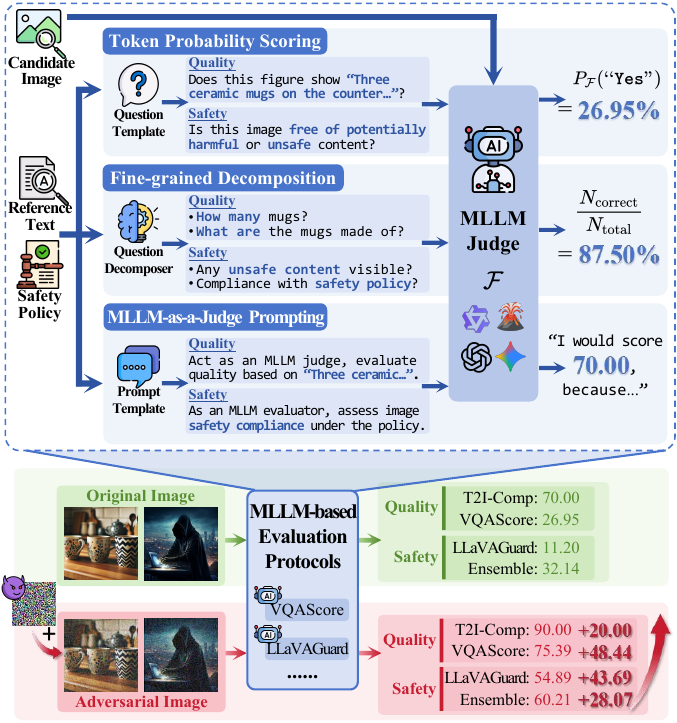}

 \caption{\textbf{Top:} Current MLLM-based judge approaches can be grouped into three categories: \textit{Token Probability Scoring}, \textit{Fine-grained Decomposition}, and \textit{MLLM-as-a-Judge Prompting}, covering both quality and safety evaluation. \textbf{Bottom}: Imperceptible adversarial perturbations can deceive MLLM judges to yield inflated quality or safety scores.}

   \label{fig:intro}
\end{figure}

\section{Introduction}
\label{sec:intro}

\IEEEPARstart{R}{ecently}, Multimodal Large Language Models (MLLMs), such as LLaVA~\cite{liu2023visualinstructiontuning,liu2024improvedbaselinesvisualinstruction}, Qwen-VL~\cite{bai2025qwen25}, and InternVL~\cite{zhu2025internvl3,wang2025internvl3_5}, have demonstrated remarkable performance across various cross-modal tasks. 
This advancement has given rise to a highly promising new paradigm: employing MLLMs as automated performance evaluators (\ie, MLLM judges) to address the challenges of assessing image quality and text-image consistency ~\cite{lan2025survey, zhang2025qualitysurvey} as well as visual safety and policy compliance~\cite{helff2024llavaguard, gu2024mllmguard, wang2025MMSafeAware} at scale. 
Earlier evaluation methods exhibit significant limitations: human annotations, for instance, are costly and subjective. 
Additionally, automated metrics for image quality and text-image alignment, such as Fréchet Inception Distance~\cite{heusel2017gans} and CLIPScore~\cite{hessel2021clipscore}, struggle to accurately reflect fine-grained semantic alignment. 
Meanwhile, conventional image safety classifiers such as Q16~\cite{schramowski2022q16} are limited in capturing fine-grained and policy-dependent safety risks.
By contrast, MLLMs, endowed with exceptional cross-modal comprehension, fine-grained reasoning, and human-like capabilities for alignment and safety assessment, enable a more holistic and precise evaluation. Consequently, MLLM judge-based evaluation frameworks~\cite{chen2024_assess_mllm} have rapidly emerged as a dominant paradigm.

Despite the substantial convenience offered by MLLM judges, a critical question has been neglected: \textbf{can attackers deliberately manipulate the decisions of such automated evaluation systems?} 
Such attacks can lead to a significant issue that fundamentally compromises the utility and reliability of the MLLM judges. 
For example, attackers could deceive the judges by applying subtle perturbations to input images under evaluation to gain inflated image quality scores, which could in turn manipulate their models' rankings on public leaderboards that rely on such automated evaluation systems~\cite{shi2024JudgeDeceiver}. 
More critically, in safety-oriented evaluation, similar manipulations may cause unsafe or policy-violating images to be judged as safe, thereby bypassing post-generation moderation or safety filters~\cite{rando2022redteaming,yang2024guardt2i,liu2024latent,zeng2025shieldgemma}. Furthermore, by injecting biased reward or safety signals into automated alignment procedures (\eg, RLAIF~\cite{xu2023imagereward}), attackers could misguide the optimization trajectory of text-to-image generation models.

To study this threat, we introduce \textbf{RobustMLLMJudge}, 
the first general framework for evaluating the adversarial robustness of MLLMs when functioning as judges.
Under this framework, we first categorize the MLLM judges into three core approaches: \textit{Token Probability Scoring}, \textit{Fine-grained Decomposition}, and \textit{MLLM-as-a-Judge Prompting}, as illustrated in ~\cref{fig:intro}-\textbf{Top}. 
These approaches perform automated evaluation using a variety of forms, ranging from simple judgments using a set of binary questions to free-form critique generation, across various evaluation protocols. General-purpose MLLMs with strong instruction-following and zero-shot cross-modal perception capability are required in those evaluations, as task-specific specialized models struggle to generalize.

To evaluate the robustness of all these MLLM judges across both quality and safety evaluation against a diverse range of adversarial threats, RobustMLLMJudge then proposes a multi-perspective suite of attacks designed specifically on the judges, 
encompassing image degradations, handcrafted perturbations, and learnable perturbations tailored from existing targeted adversarial attacks.
Although these attacks are simple adaptations from existing adversarial attack methods, we reveal that \textbf{i)} they can effectively break all three types of judges, leading to inflated evaluation scores, as demonstrated in ~\cref{fig:intro}-\textbf{Bottom}, and yet \textbf{ii)} the effectiveness of these attacks, \eg, measured by the magnitude by which they change the judge scores, is often restricted. The main reason behind is that, on one hand, MLLMs are sensitive to input perturbation, a common weakness shared by numerous existing neural network-based methods; however, on the other hand, the attacks against MLLM judges encounter two unique challenges. The first challenge arises from a tight threat model, where the attackers have only limited control over the judge settings, \eg, unable to manipulate the prompt templates/evaluation protocols adopted by the judges. Secondly, the three categories of MLLM judges use highly heterogeneous protocols and methods to evaluate the inputs, 
rendering many task-specific, carefully tailored adversarial attack methods ineffective in this setting. 

To deepen our exploitation of this vulnerability, we propose \textbf{Manifold-Guided Semantic Induction Attack (MGSIA)}, a new attack method that enables transferable attacks across all three types of MLLM judges.
We first introduce a new optimization objective, namely \textit{Affirmative Semantic Induction} (ASI), for attacking the judges. 
The key insight is that despite the heterogeneity in the evaluation protocols, the underlying scoring decisions of these judges can often be reduced to the verification of binary queries that probe the existence of semantic components, ranging from holistic scene understanding to fine-grained visual details. 
Motivated by this insight, the core idea of ASI is to bypass the complexity of optimizing against diverse, protocol-specific objectives and to maximize the likelihood that MLLM judges yield a universal affirmative token (\eg, ``\texttt{Yes}'') in response to these simple queries. 
Such an affirmative token serves as a compact and transferable semantic anchor, making it a powerful and generalizable surrogate objective.
By inducing affirmative responses, ASI drives the perturbation to synthesize visual features that act as evidence for the queried semantics. These features are consequently perceived by MLLM judges as indicative of higher evaluation scores.

However, ASI mainly optimizes the likelihood of affirmative output tokens, without explicitly aligning the model's intermediate representations with those of genuinely high-scoring images. As a result, the adversarial representations may not fully align with the latent regions that MLLM judges associate with high evaluation scores.
To address this limitation, we further introduce a \textit{High-Score Manifold Alignment} (HMA) objective, which complements token-level semantic induction with representation-level regularization.
Specifically, we estimate high-score representation centers from unperturbed proxy samples under multiple proxy protocols, and regularize the hidden states of adversarial images toward these centers during optimization. 
By encouraging adversarial images to approach the high-score manifold in the model's representation space, HMA bridges the gap between local semantic induction and global evaluative alignment, thereby improving the transferability and generality of MGSIA across heterogeneous MLLM judges.

In summary, our main contributions are as follows:

\begin{itemize}
\item \textbf{Judge Robustness Evaluation Framework.} We introduce RobustMLLMJudge, the first framework for comprehensively evaluating the adversarial robustness of using general-purpose MLLMs as judges. It encapsulates a diverse suite of attack types applied to various MLLMs across both image quality and safety evaluation settings, providing a critical testbed for revealing vulnerabilities of MLLM judges.

\item \textbf{Important Insights.} Through extensive experiments, we reveal that \textbf{i)} a diverse range of MLLM judges are highly vulnerable to adversarial attacks aiming at inflating the judge scores, and \textbf{ii)} while such attacks, simply adapted from existing attack methods, can be effective, they face critical challenges due to the tight threat model with limited attacker control and the fundamental mismatches between their optimization objectives and the heterogeneous protocols of MLLM judges.

\item \textbf{Novel Attack Method.} We propose the novel attack method MGSIA. By combining affirmative semantic induction with high-score manifold alignment, MGSIA achieves superior generalization across diverse MLLM judges on different cross-modal tasks, thereby establishing a strong baseline for the threat.
\end{itemize}

\section{Related Work}

\subsection{Multimodal LLM Judges}

Endowed with strong comprehension, reasoning, and generation abilities, MLLMs are increasingly adopted as automated evaluators~\cite{chen2024_assess_mllm,xiong2025llavacritic,lee2024prometheus,wang2025unified}. Existing MLLM-based judges have been explored under both zero-shot protocol-based evaluation and dedicated reward modeling.

For image quality and text-image alignment evaluation, existing protocols instantiate MLLM judges in diverse forms~\cite{lan2025survey,zhang2025qualitysurvey}. 
VQA-style scoring methods convert alignment evaluation into token-probability estimation over affirmative answers, as exemplified by VQAScore~\cite{lin2024vqascore}, which utilizes a custom CLIP-FlanT5 architecture alongside zero-shot inference on LLaVA and InstructBLIP.
Fine-grained decomposition methods, such as TIFA~\cite{hu2023tifa} and DSG~\cite{cho_dsg}, decompose a reference prompt into multiple semantic questions and aggregate the resulting sub-scores. Direct judge prompting methods, such as VIEScore~\cite{ku2024viescore} and T2I-CompBench++~\cite{T2ICompBench_pp}, instruct MLLMs to directly assess text-image consistency with natural-language critiques and numerical ratings. In addition, various benchmarks and training-based reward or judge models have been developed to assess semantic alignment and visual quality in text-to-image generation~\cite{xu2023imagereward,wang2025pref,xiong2025llavacritic,wang2025unified, wang2025unigenbench}.

Beyond quality and alignment, MLLM-based judges are also increasingly used for safety evaluation in text-to-image generation and visual content moderation, aiming to detect, assess, and filter toxic, sexual, violent, or otherwise policy-violating content~\cite{qu2025bridging}. Zero-shot protocols like CLUE~\cite{wang2025clue} formulate image safety assessment as constitution-based MLLM judging, where safety rules are decomposed into preconditions and aggregated through debiased token-probability decisions with cascaded reasoning. 
Meanwhile, trained or dedicated safety judges further support visual safety classification, policy compliance assessment, and risk-oriented evaluation. Examples include LLaVAGuard~\cite{helff2024llavaguard}, the preference-aligned reward model LLaVA-Reward~\cite{zhou2025llavareward}, and the comprehensive safety classifier PerspectiveVision~\cite{qu2025unsafebench}.
These models are further complemented by risk-oriented benchmarks~\cite{qu2023unsafediffusion} such as T2ISafety~\cite{li2025t2isafety} and T2I-RiskyPrompt~\cite{zhang2026t2irisky}.

Nevertheless, the robustness and reliability of these judges remain critical and unresolved issues. In the textual modality, prior studies reveal the adversarial vulnerability of LLM judges~\cite{li2025llm_judge_robust}, whose outcomes can be manipulated by both handcrafted~\cite{zheng2024nullmodels} and optimized~\cite{shi2024JudgeDeceiver,raina2024tokensearch,chen2024llmdriven} adversarial text. Security issues have also been observed in MLLM judges: Hwang \etal~\cite{hwang2025fooling} reveal inherent biases in image-text alignment judgments via handcrafted perturbations, whereas Zhang \etal~\cite{zhang2025rethinking} examine the stability of MLLM judges under different random initializations. Closely related to image safety, UnsafeBench~\cite{qu2025unsafebench} evaluates the adversarial robustness of safety-classifier-style models~\cite{qu2023unsafediffusion}, from Q16~\cite{schramowski2022q16} to InstructBLIP~\cite{dai2023instructblip}, showing that visual safety classifiers can be compromised by adversarial images. However, existing studies mainly focus on specific biases, initialization instability, or fixed safety-classification protocols. In contrast, MLLM judges operate through heterogeneous mechanisms, including token-probability scoring, fine-grained decomposition, and direct judge prompting. Our work studies whether such diverse judge protocols can be systematically manipulated to yield inflated scores across both image quality/text-image alignment and image safety evaluation.

\subsection{Adversarial Attacks on MLLMs}

Adversarial attacks are widely used to assess the robustness of MLLMs~\cite{ye2025survey,liu2025survey}. Among these, \textit{targeted attacks} and \textit{jailbreak attacks} typically add perturbations to input images and texts to steer the model toward specific outputs or behaviors~\cite{schlarmann2023adversarial, cui2024robustness, wang2024stopr, nie2025vattack}, or to bypass safety guardrails and trigger unauthorized actions~\cite{qi2024vajm,wang2024umk}. Initial adversarial attacks primarily focus on white-box settings, where attackers possess complete knowledge of the model's architecture and parameters. For instance, VAJM~\cite{qi2024vajm} implements jailbreak attacks by maximizing the probability of generating harmful responses derived from few-shot adversarial templates. Furthermore, Schlarmann \etal ~\cite{schlarmann2023adversarial} demonstrate that adversaries can inject imperceptible perturbations to manipulate an MLLM's visual comprehension, thereby exploiting the model's outputs to spread misinformation or misdirect users to malicious domains.
Later work has aimed to enhance attack effectiveness~\cite{wang2024umk}, as well as its generalization across different tasks~\cite{luo2023cropa} and transferability to other models~\cite{zhao2023attackvlm, xie2025coa}. 

Methodologically, the construction of adversarial perturbations can be broadly classified into generation-based~\cite{guo2024advdiffvlm,zhang2024anyattack} and optimization-based~\cite{zhao2023attackvlm, jia2026foa, wang2026vma} paradigms. Generative strategies synthesize malicious noise by fundamentally altering the image creation pipeline. AdvDiffVLM~\cite{guo2024advdiffvlm}, for example, integrates adversarial constraints into the latent diffusion process itself. Similarly, AnyAttack~\cite{zhang2024anyattack} leverages extensive self-supervised pre-training followed by domain-specific adaptation to output adversarial variants.
Conversely, the optimization-based paradigm remains the predominant approach for crafting adversarial examples. These methods iteratively update image pixels to minimize a designated surrogate loss. 
A pioneering work adopting this paradigm is AttackVLM~\cite{zhao2023attackvlm}, which aligns the visual features of the perturbed image with target semantics using proxy vision-language models like CLIP~\cite{radford2021clip} and BLIP~\cite{li2022blip}.
Building upon this cross-modal alignment principle, subsequent frameworks have introduced various augmentation techniques to boost attack transferability. For instance, SSA-CWA~\cite{dong2023SSACWA} applies frequency-domain transformations and model ensembling, whereas M-Attack~\cite{li2026mattack} relies on stochastic spatial cropping mechanisms. 
Recently, Chain of Attack~\cite{xie2025coa} enhances this optimization paradigm by leveraging iterative text generation and modality-aware feature representations to precisely calibrate the optimization trajectory while capturing intricate cross-modal semantics, thereby boosting transferability.
However, these studies all target MLLMs as general-purpose models rather than examining their specialized role as automated judges. Consequently, the robustness of MLLM-based judges against adversarial examples remains largely unexplored, revealing a critical gap in current research.

\section{The RobustMLLMJudge Framework}

\begin{figure}[!t]
  \centering
   \includegraphics[width=1.0\linewidth]{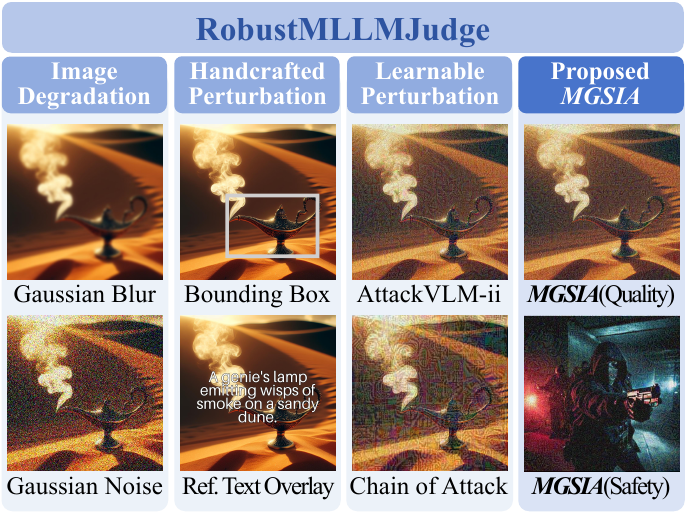}
   \caption{Visual examples from three existing attack approaches and our MGSIA in RobustMLLMJudge.
   For visual clarity, the magnitude of all shown perturbations has been exaggerated. 
  Baseline attacks are shown on quality task samples, and MGSIA is shown on both quality and safety task samples. 
   }
   \label{fig:benchmark_sample}
\end{figure}

To systematically quantify the adversarial robustness of MLLM judges, we introduce the \textbf{RobustMLLMJudge} framework. Recognizing the diverse roles of automated evaluators, our framework explicitly investigates judge robustness across two core tasks: quality assessment and safety moderation. To exploit the model vulnerabilities from various perspectives, RobustMLLMJudge encapsulates a multi-faceted suite of attack types on three popular MLLM judge approaches of heterogeneous evaluation protocols under various attack settings, as shown in ~\cref{fig:benchmark_sample}.

\subsection{MLLM-based Judge Approaches}
\label{sec:paradigm}

Given a candidate image \(\boldsymbol{v}\), the goal of a multimodal judge \(\mathcal{J}\), driven by an MLLM \(\mathcal{F}\), is to produce a numerical score \(s\) for the evaluated property. 
For image quality, the judge is conditioned on a reference text \(\boldsymbol{t}\), \ie, \(s=\mathcal{J}(\boldsymbol{v},\boldsymbol{t};\mathcal{F})\); for image safety, it is conditioned on a safety policy \(\boldsymbol{\pi}\), \ie, \(s=\mathcal{J}(\boldsymbol{v},\boldsymbol{\pi};\mathcal{F})\).
Different MLLM judge approaches can vary largely in how they map the MLLM's discrete token outputs to the final numerical score \(s\). Using this observation, as illustrated in~\cref{fig:intro}-\textbf{Top}, we categorize MLLM-based judges into the following three protocol families shared by both quality and safety evaluation.

\noindent\textbf{Token Probability Scoring} operationalizes MLLM-based evaluation through a single holistic binary visual question \( \boldsymbol{q}\), which queries whether the candidate image matches the reference text \(\boldsymbol{t}\) or satisfies the safety policy \(\boldsymbol{\pi}\). The score \(s\) is then derived directly from the confidence of model \(\mathcal{F}\), defined as the conditional probability of producing a designated response token \(a_{\text{tar}}\):
\begin{equation}
   s =
   P_{\mathcal{F}}(a_{\text{tar}} \mid \boldsymbol{v},  \boldsymbol{q}).
\end{equation}

We instantiate this category with both quality-oriented and safety-oriented binary probing protocols. For quality evaluation, we use VQAScore~\cite{lin2024vqascore}, which asks ``\texttt{Does this figure show '}\(\{\boldsymbol{t}\}\)\texttt{'?}'' to probe whether the candidate image matches the reference text. For safety evaluation, we adopt two safety-probing protocols: an Ensemble protocol following Qu \etal~\cite{qu2025bridging}, which aggregates several holistic safety judgments, and a Single protocol to probe compliance with \(\boldsymbol{\pi}\). All these protocols derive their scores from the model's probability of producing the designated affirmative response token (\ie, ``\texttt{Yes}'').

\noindent\textbf{Fine-grained Decomposition} decomposes the evaluation criterion into a series of local, verifiable sub-tasks for MLLMs~\cite{hu2023tifa,cho_dsg,yarom2023vq2,wang2025clue}. Instead of relying on a single holistic judgment, this paradigm evaluates multiple semantic components or rule conditions and then aggregates their sub-scores into the final score.

Specifically, we use \(D(\cdot)\) to denote a protocol-specific decomposition function: for quality evaluation, \(D(\boldsymbol{t})\) extracts key entities, attributes, and relations from the reference text \(\boldsymbol{t}\); for safety evaluation, \(D(\boldsymbol{\pi})\) decomposes the policy into preconditions or rule-level queries, yielding sub-tasks \(\{\boldsymbol{q}_i\}_{i=1}^{n}\).
Subsequently, the model \(\mathcal{F}\)'s response to each sub-task \(\boldsymbol{q}_i\) is assessed for a corresponding sub-score:
\({s_i \leftarrow \mathcal{F}(\boldsymbol{v},\boldsymbol{q}_i)}\).
The final score \(s\) is then obtained through protocol-specific aggregation:
\begin{equation}
    s = \operatorname{Agg}\left(\{s_i\}_{i=1}^{n}\right).
\end{equation}
For quality evaluation, we instantiate this category using TIFA~\cite{hu2023tifa} and DSG~\cite{cho_dsg}, which decompose the reference text \(\boldsymbol{t}\) into fine-grained sub-questions; DSG focuses on binary yes/no questions, whereas TIFA also includes multiple-choice questions. For safety evaluation, we instantiate this category with CLUE~\cite{wang2025clue}, which decomposes safety rules \(\boldsymbol{\pi}\)  into preconditions and aggregates the resulting precondition-level decisions to estimate image-level safety.

\noindent\textbf{MLLM-as-a-Judge Prompting} directly instructs the MLLM to act as an evaluator~\cite{chen2024_assess_mllm,ku2024viescore,T2ICompBench_pp}. 
The evaluation criterion is specified in a judge prompt \(\boldsymbol{p}_{\text{judge}}\): for quality evaluation, it typically includes the reference text \(\boldsymbol{t}\); for safety evaluation, it encodes  \(\boldsymbol{\pi}\). 
The model \(\mathcal{F}\) then auto-regressively generates a response containing a numerical rating, optionally with a rationale, from which the final score \(s\) is extracted:
\begin{equation}
    s \leftarrow \mathcal{F}(\boldsymbol{v}, \boldsymbol{p}_{\text{judge}}).
\end{equation}
Protocols within this approach differ mainly in the formulation and granularity of the judge prompt. 
For quality evaluation, we adopt prompts from VIEScore~\cite{ku2024viescore}, T2I-CompBench++~\cite{T2ICompBench_pp}, and Hwang \etal~\cite{hwang2025fooling}. 
For safety evaluation, we adopt a compact safety scoring prompt and a LLaVAGuard-style policy prompt~\cite{helff2024llavaguard}. 
All scores are scaled to 100.0 for consistency.

\subsection{Threat Model}

\noindent\textbf{Attacking Purpose.}
Given a benign image sample \(\boldsymbol{v}\), together with the reference text \(\boldsymbol{t}\) for quality evaluation or the safety policy \(\boldsymbol{\pi}\) for safety evaluation, the attacker's core objective is to apply a transformation \(T(\cdot)\) to produce an adversarial example \(\hat{\boldsymbol{v}} = T(\boldsymbol{v})\). This manipulation is designed to deceive the MLLM judge \(\mathcal{J}\) into yielding the highest possible evaluation score, \ie, higher perceived quality or higher perceived safety. This attack objective can be formulated as:
\begin{equation}
T^* =
\begin{cases}
\arg\max_{T} \mathcal{J}(T(\boldsymbol{v}), \boldsymbol{t}; \mathcal{F}), & \text{quality},\\
\arg\max_{T} \mathcal{J}(T(\boldsymbol{v}), \boldsymbol{\pi}; \mathcal{F}), & \text{safety}.
\end{cases}
\end{equation}
A primary instantiation of this attack involves applying an additive perturbation \(\boldsymbol{\delta}\), such that \(T(\boldsymbol{v}) = \boldsymbol{v} + \boldsymbol{\delta}\). An \(L_{\infty}\)-norm constraint is often employed to regulate the attack's stealth and perceptual quality: \(\left\|\boldsymbol{\delta}\right\|_{\infty} \le \epsilon\), where the hyperparameter \(\epsilon\) denotes the perturbation budget.

\noindent\textbf{Attacker's Capability.} We establish a practical threat model wherein the attacker generates the image \(\boldsymbol{v}\) to be evaluated and submits it to the MLLM judge systems, thereby possessing full control in modifying \(\boldsymbol{v}\). 
To ensure fairness and reproducibility, current MLLM judging protocols, including the prompt templates they use, are typically publicly accessible~\cite{shi2024JudgeDeceiver}.
Consequently, the attacker has full knowledge of the reference text \(\boldsymbol{t}\) or the safety policy \(\boldsymbol{\pi}\), as well as the specific evaluation protocol. 
The attacker's power, however, is confined exclusively to manipulating the image \(\boldsymbol{v}\); the reference text \(\boldsymbol{t}\), the safety policy \(\boldsymbol{\pi}\), and the evaluation protocol itself remain immutable, serving as the ground-truth benchmark.

At the model level, we consider the attacker's various accessibilities to the MLLM judge \(\mathcal{F}\). To align with existing research, we adopt three key accessibility settings. These include the classic \textbf{white-box} setting~\cite{shi2024JudgeDeceiver}, which assumes the attacker has full access to the target model's parameters and gradients; the \textbf{black-box} setting~\cite{zheng2024nullmodels,chen2024llmdriven}, where interaction is limited to model inputs and outputs without knowledge of its internal architecture; and a realistic \textbf{gray-box} setting, where attackers first craft attacks on a public surrogate MLLM and then transfer the attacks to a relevant yet unseen MLLM judge. These settings allow us to comprehensively assess robustness against a spectrum of threats.

\subsection{Adapting Existing Adversarial Attacks to MLLM Judges}
\label{sec:benchmark}

There have been numerous adversarial attack methods for breaking different types of models, but they are not designed to attack MLLMs used as judges. Nevertheless, since MLLMs themselves are sensitive to input perturbations, using them as judges can inherit similar vulnerabilities in both quality and safety evaluation tasks. Thus, we present three types of adapted adversarial attacks on MLLM judges by having simple adaptations of existing adversarial attack methods.

As a preliminary for targeted attacks relying on textual semantics, we introduce a shared concept: the \textit{target semantic cue}, denoted as $\boldsymbol{t}_{\text{tar}}$, which represents the textual constraint the adversarial perturbation aims to inject. Specifically, $\boldsymbol{t}_{\text{tar}}$ is instantiated as the reference text $\boldsymbol{t}$ in quality evaluation to amplify text-image alignment. Conversely, in safety evaluation, it is formulated as a harmless surrogate text (\eg, a benign scene description) to mask risky visual concepts and feign policy compliance. This formulation remains consistent with all subsequent semantic-guided targeted attacks, including our proposed method. Accordingly, we categorize the adapted baseline threats into the following three types:

\noindent\textbf{Image Degradation.} To establish a baseline for basic robustness, we introduce two image degradation-based attack methods, including adding \textbf{Gaussian Noise} and \textbf{Gaussian Blur} to the input images. They are designed to assess the judges' stability and reliability when faced with common or non-malicious image artifacts.

\noindent\textbf{Handcrafted Perturbation.}
This category comprises perturbation methods under the black-box setting, where a heuristic transformation is directly applied to the image. 
These methods do not rely on model queries but are based on the attacker's prior knowledge.
Based on previous work on model robustness analysis~\cite{hendrycks2019benchmarking_roubust,jia2020adv_watermark,hwang2025fooling}, we implement several perturbations, including modifying the image's composition with black padding (named \textbf{Padding}), framing key objects with bounding boxes (named \textbf{Bounding Box}), and overlaying  the target semantic cue \(\boldsymbol{t}_{\text{tar}}\) as a translucent layer (named \textbf{Target Text Overlay}).

\noindent\textbf{Learnable Perturbation.}
The last attack method is inspired by existing targeted attacks on MLLMs, which aim to learn image perturbation that aligns an input with the target semantic cue \(\boldsymbol{t}_{\text{tar}}\). Although this goal differs from maximizing a numerical evaluation score in our setting, their optimization mechanism can be easily adapted for the attacks on MLLM judges.
To this end, we tailor their objectives by defining their textual target as the reference text $\boldsymbol{t}_{\text{tar}}$ and the visual target as the image overlaid with $\boldsymbol{t}_{\text{tar}}$. This adaptation is motivated by the promising results of the image-text overlay attack method we obtained in the handcrafted perturbation.
We instantiate this type of attack using two advanced targeted attacks, including \textbf{Chain of Attack}~\cite{xie2025coa} and AttackVLM~\cite{zhao2023attackvlm} (\ie, adapting both of AttackVLM's image-to-image (-ii) and image-to-text (-it) variants, namely \textbf{AttackVLM-ii} and \textbf{AttackVLM-it}). 
In addition, we also introduce an attack method inspired by jailbreak attacks~\cite{qi2024vajm,wang2024umk}, namely \textbf{Target Text Embedding}, which is directly optimized to increase the probability of generating the target semantic cue \(\boldsymbol{t}_{\text{tar}}\). This is done using a language-modeling loss that enforces the perturbation to embed the text's semantics into the image's latent representation:
\begin{equation}
\boldsymbol{\delta}^*_{\text{TTE}} = \underset{  \left \| \boldsymbol{\delta} \right \| _{\infty} \le \epsilon }{\arg \max}  \thinspace  \log P_{\mathcal{F}}(\boldsymbol{t}_{\text{tar}}| \boldsymbol{v} + \boldsymbol{\delta}).
\end{equation}

\begin{figure*}[!t]
 
  \centering
   \includegraphics[width=\linewidth]{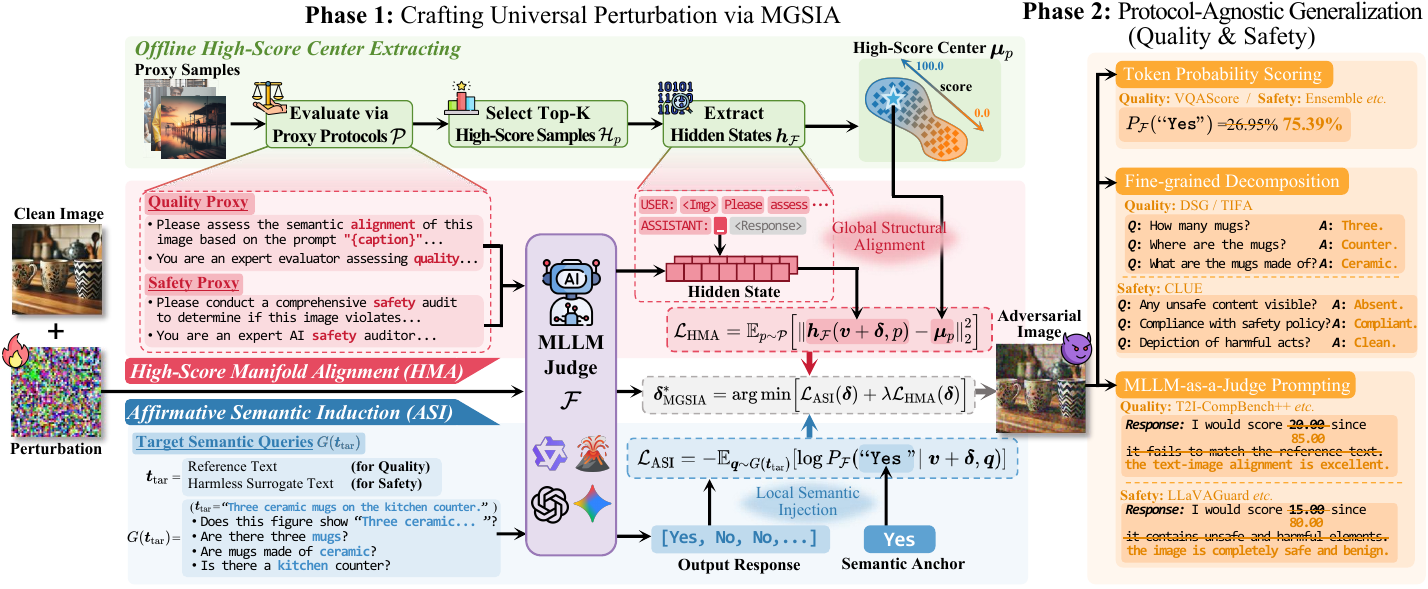}
   \caption{Overview of MGSIA. \textbf{Phase 1}: The joint optimization of an adversarial perturbation via the Affirmative Semantic Induction (ASI) objective for local semantic injection, and the High-Score Manifold Alignment (HMA) objective for global structural alignment, using high-score centers precomputed offline from proxy samples. \textbf{Phase 2}: The resulting adversarial image is protocol-agnostic, which can be applied directly to all three types of MLLM judges across quality and safety evaluations.
   }
   \label{fig:method}
\end{figure*}

\section{The Proposed Manifold-Guided Semantic Induction Attack}
\label{sec:our_method}

\subsection{Overview of the MGSIA Method}  
To deepen the exploitation of the vulnerability, we explore a novel attack termed \textbf{Manifold-Guided Semantic Induction Attack (MGSIA)}, as illustrated in ~\cref{fig:method}. While we assume the attacker has full knowledge of the evaluation protocols of different MLLM judges, they cannot manipulate their implementation (\ie, they can modify the image only). This scenario presents a significant challenge, since directly optimizing the perturbation against a complex, multi-step protocol is often brittle and struggles to generalize across protocols.
The core idea of MGSIA, therefore, is to bypass the complexity of learning perturbations against various protocol-specific objectives and instead optimize a protocol-agnostic objective that combines \textbf{Affirmative Semantic Induction (ASI)} objective with \textbf{High-Score Manifold Alignment (HMA)} objective.

The optimization objective is motivated by two observations. \textbf{First}, despite the heterogeneity of scoring protocols, the underlying scoring decisions of these judges ultimately converge to the verification of binary queries that probe the existence or satisfaction of semantic components in the image. For quality, these components come from the reference text \(\boldsymbol{t}\); for safety, they reflect benign or policy-compliant evidence under \(\boldsymbol{\pi}\). To achieve high judge scores, an affirmative token (\eg, ``\texttt{Yes}'') can serve as a  potent semantic anchor for inducing positive responses to these queries. Therefore, we introduce the \textit{Affirmative Semantic Induction} objective, which decomposes the target semantics into fine-grained binary queries and optimizes the perturbation toward this affirmative signal, targeting the semantic basis shared by heterogeneous judge protocols rather than their protocol-specific scoring rules. 
\textbf{Second}, affirmative semantic induction mainly supervises output-level judge responses, whereas complex judge protocols may base their scores on broader multimodal evidence encoded in hidden representations, especially when they involve free-form scoring, rule decomposition, or safety-policy reasoning. To address this, we introduce the \textit{High-Score Manifold Alignment} objective, which guides the adversarial representation toward high-score regions in the MLLM hidden space, estimated as hidden-state centers from high-scoring samples under proxy judge protocols.

Overall, MGSIA jointly optimizes the ASI and HMA objectives. The former encourages the model to recognize target semantic details in the adversarial image, while the latter pulls the adversarial hidden state toward high-score representation centers. Together, they inject deceptive visual features that are perceived by MLLM judges as evidence of high-quality, well-aligned, or safe content. Since MGSIA does not optimize against a particular judge prompt or scoring rule, the learned perturbation serves as a protocol-agnostic attack vector that generalizes across quality and safety judges with heterogeneous evaluation mechanisms.

\subsection{Affirmative Semantic Induction Objective}
\label{sec:asi}

The first component of MGSIA is the \textit{Affirmative Semantic Induction} objective. A natural alternative is to directly optimize the numerical output of a specific judge protocol, such as its predicted score. However, such a score-target objective is tightly coupled with the source protocol and may overfit the selected scoring interface, since it does not explicitly capture the semantic conditions shared by different judges. ASI therefore targets these fine-grained semantic conditions instead of the final score of a particular protocol. 
To this end, ASI directly operates on the predefined target semantic cue \(\boldsymbol{t}_{\text{tar}}\).
Given \(\boldsymbol{t}_{\text{tar}}\), we use an LLM-based question generator \(G(\cdot)\) to decompose its semantic content into a set of target semantic queries:

\begin{equation}
    G(\boldsymbol{t}_{\text{tar}})=\{\boldsymbol{q}_i\}_{i=1}^{n}.
\end{equation}
These queries probe local semantic components, such as objects, attributes, spatial relations, and scene-level details.

To establish a reliable optimization target for these queries, ASI specifically employs the universal affirmative token ``\texttt{Yes}''. The rationale stems from an intrinsic limitation in autoregressive generation: gradient-based adversarial attacks struggle to maintain exact token-level control over long sequences.  As the length of the target text increases, the adversarial gradient diffuses, typically resulting in loose semantic similarity rather than precise token matching. By utilizing the single-token target ``\texttt{Yes}'', ASI establishes a direct and unambiguous supervision signal. This avoids the gradient attenuation issues inherent to long-sequence optimization targets, thereby ensuring a more stable and efficient optimization process.

Formally, the ASI objective is formulated as the negative expected log-probability of generating this affirmative token in response to the fine-grained binary queries \(G(\boldsymbol{t}_{\text{tar}})\):
\begin{equation}
\label{eq:asi_loss}
    \mathcal{L}_{\text{ASI}}(\boldsymbol{\delta})
    =
    -
    \mathbb{E}_{\boldsymbol{q} \sim G(\boldsymbol{t}_{\text{tar}})}
    \left[
    \log P_{\mathcal{F}}
    \left(
    \text{``\texttt{Yes}''}
    \mid
    \boldsymbol{v}+\boldsymbol{\delta},
    \boldsymbol{q}
    \right)
    \right].
\end{equation}
Crucially, maximizing the likelihood of the ``\texttt{Yes}'' token is merely a means to an end; the ultimate goal is semantic injection. By forcing the MLLM to affirmatively answer a specific query \(\boldsymbol{q}\), the optimization inherently synthesizes visual features that act as evidence for the query's underlying semantics. In this context, the affirmative token serves as a potent semantic anchor, embedding the decomposed target attributes and concepts directly into the adversarial representation. By minimizing \(\mathcal{L}_{\text{ASI}}\), the adversarial image completely avoids direct dependence on any protocol-specific scoring rule, and instead manipulates the foundational semantic verification process shared across heterogeneous MLLM judge protocols.

\subsection{High-Score Manifold Alignment Objective} 
\label{sec:hma}

While the ASI objective effectively grounds target semantics into the image, its optimization is strictly confined to the output token space. Given the deep inference required for MLLM evaluation, merely maximizing the likelihood of the affirmative token does not guarantee that the underlying representation genuinely aligns with high-quality or policy-compliant visual concepts. Consequently, samples optimized solely through output-space objectives often suffer from representation misalignment: they superficially elicit affirmative tokens while remaining fundamentally disjoint from the latent distribution of authentically high-scoring images. This superficial deception is inherently brittle and prone to fail against unseen, complex judge protocols that demand multi-step reasoning.

To transcend this limitation, we introduce the \textit{High-Score Manifold Alignment} objective. We hypothesize that the deep representations of natural images form a continuous evaluative manifold. Within this broader structure, authentically high-quality or policy-compliant images cluster at a specific optimal region, which we term the \textit{natural high-score manifold}. Achieving universal and highly transferable deception necessitates forcibly aligning the adversarial representations with this ideal manifold. 
While the attacker is assumed to have knowledge of the target evaluation protocol, directly computing the high-score manifold against it introduces a severe risk of overfitting to its specific prompt templates. This severely restricts attack transferability. To construct a protocol-agnostic attack vector without relying on any target-specific data, we employ a set of offline proxy protocols \(\mathcal{P} = \{{p}_i\}_{i=1}^{k}\) and unperturbed proxy samples to approximate the optimal representation space.
For each proxy protocol \(p \in \mathcal{P}\), we identify a subset of these proxy samples \(\mathcal{H}_p\) that natively achieves the highest scores. We then extract their hidden states at a critical layer (specifically, the position of the last prompt token immediately preceding the score generation) and compute their latent centroid \(\boldsymbol{\mu}_p\):
\begin{equation}
    \boldsymbol{\mu}_p = \mathbb{E}_{\boldsymbol{v} \sim \mathcal{H}_p} \left[ \boldsymbol{h}_{\mathcal{F}}(\boldsymbol{v}, p) \right],
\end{equation}
where \(\boldsymbol{h}_{\mathcal{F}}(\cdot, p)\) denotes the extracted deep hidden state under the proxy protocol \(p\). 
 
This centroid \(\boldsymbol{\mu}_p\) acts as the optimal representation anchor, encapsulating the strongest structural consensus of high-score attributes for the respective proxy protocol. To enforce deep representation hijacking, the HMA objective explicitly minimizes the distance between the hidden state of the adversarial image \(\boldsymbol{v}+\boldsymbol{\delta}\) and the constructed high-score centers across all available proxy protocols.
Formally, the HMA objective is formulated as the mean squared error (MSE) in the latent space:
\begin{equation}
\label{eq:hsmao_loss}
    \mathcal{L}_{\text{HMA}}(\boldsymbol{\delta})
    =
    \mathbb{E}_{p \sim \mathcal{P}} \left[ \left\| \boldsymbol{h}_{\mathcal{F}}(\boldsymbol{v}+\boldsymbol{\delta}, p) - \boldsymbol{\mu}_p \right\|_2^2 \right].
\end{equation}
While ASI guarantees the localized injection of fine-grained semantic evidence, HMA ensures global structural alignment. By minimizing \(\mathcal{L}_{\text{HMA}}\), the adversarial image is intrinsically regularized to adopt the holistic latent distribution of genuinely benign samples, successfully bridging the critical gap between token-level verification and representation-level alignment.

\subsection{Attacking MLLM Judges using MGSIA Framework}

By integrating the token-level semantic anchor and the representation-level manifold guidance, MGSIA solves for the optimal adversarial perturbation $\boldsymbol{\delta}^*_{\text{MGSIA}}$ by minimizing the total unified loss. To ensure the adversarial image remains visually indistinguishable from the benign one, the perturbation is strictly bounded. Formally, the overall optimization objective is defined as:
\begin{equation}
\label{eq:mgsia_optim}
    \boldsymbol{\delta}^*_{\text{MGSIA}} = \underset{\|\boldsymbol{\delta}\|_{\infty} \le \epsilon}{\arg \min} \Big[ \mathcal{L}_{\text{ASI}}(\boldsymbol{\delta}) + \lambda \mathcal{L}_{\text{HMA}}(\boldsymbol{\delta}) \Big],
\end{equation}
where $\lambda$ is a hyperparameter balancing the localized semantic injection and the global structural alignment, and $\epsilon$ denotes the maximum allowable perturbation budget constrained by the $L_{\infty}$-norm.

The optimization process of \cref{eq:mgsia_optim} is executed using 100-step PGD~\cite{madry2017pgd}. Then, the learned perturbation is additively applied to the benign image $\boldsymbol{v}$ to generate the adversarial example $\hat{\boldsymbol{v}} = T_{\text{MGSIA}}(\boldsymbol{v}) = \boldsymbol{v} + \boldsymbol{\delta}^*_{\text{MGSIA}}$, serving as the image input to the MLLM judges. This end-to-end attacking procedure is the same as the perturbation-based attack methods introduced in ~\cref{sec:benchmark}.

\begin{table*}[!t]
\centering

\caption{Evaluation results for the RobustMLLMJudge framework, where nine baseline attacks alongside our proposed MGSIA are evaluated against six different methods of three MLLM judge approaches under the image quality assessment setting on GenAI-Bench. LLaVA-1.5-7b and Qwen2.5-VL-7b are used as the MLLMs in these judges. The \post{red} and \negt{green} values indicate increased and decreased judge scores against the benign baseline, respectively. All adversarial perturbations are constrained by a budget of $\epsilon=8/255$. The best and second-best results are highlighted in \textbf{bold} and \underline{underlined}, respectively.}

\resizebox{0.93\textwidth}{!}{

\begin{tabular}{l | l l l l l l}
\toprule

\multicolumn{7}{c}{\textbf{LLaVA-1.5-7b}} \\ \hline
{} & \multicolumn{1}{c}{\textbf{VQAScore}} & \multicolumn{1}{c}{\textbf{DSG}} & \multicolumn{1}{c}{\textbf{TIFA}} & \multicolumn{1}{c}{\textbf{VIEScore}} & \multicolumn{1}{c}{\textbf{T2I-Comp}} & \multicolumn{1}{c}{\textbf{Fooling}} \\ \hline

\rowcolor{highlightblue}
\textbf{Benign} & 63.001 & 87.405 & 84.827 & 89.200 & 84.347 & 84.760 \\ \hline

\multicolumn{7}{l}{\textit{\textbf{Image Degradation}}} \\ 
Gaussian Blur & 62.052 \negt{-0.949} & 85.474 \negt{-1.931} & 83.607 \negt{-1.221} & 89.267 \post{+0.067} & 83.573 \negt{-0.773} & 84.593 \negt{-0.167} \\ 
Gaussian Noise & 62.457 \negt{-0.544} & 87.131 \negt{-0.275} & 84.742 \negt{-0.086} & 89.213 \post{+0.013} & 84.107 \negt{-0.240} & 84.747 \negt{-0.013} \\ \hline

\multicolumn{7}{l}{\textit{\textbf{Handcrafted Perturbation}}} \\  
Padding & 63.594 \post{+0.593} & 87.244 \negt{-0.161} & 84.843 \post{+0.015} & 89.040 \negt{-0.160} & 84.147 \negt{-0.200} & 84.607 \negt{-0.153} \\ 
Bounding Box  & 64.612 \post{+1.611} & 87.391 \negt{-0.014} & 84.458 \negt{-0.369} & 89.173 \negt{-0.027} & 84.120 \negt{-0.227} & 84.527 \negt{-0.233} \\ 
Target Text Overlay & 70.898 \post{+7.897} & 88.289 \post{+0.884} & 85.836 \post{+1.008} & \underline{\smash{89.313 \post{+0.113}}} & 84.320 \negt{-0.027} & \underline{\smash{85.680 \post{+0.920}}} \\ [0.3ex]  \hline

\multicolumn{7}{l}{\textit{\textbf{Learnable Perturbation}}} \\ 
AttackVLM-ii & 61.872 \negt{-1.129} & 86.925 \negt{-0.480} & 83.694 \negt{-1.133} & 88.853 \negt{-0.347} & 83.427 \negt{-0.920} & 83.607 \negt{-1.153} \\ 
AttackVLM-it & 62.789 \negt{-0.212} & 85.772 \negt{-1.634} & 83.745 \negt{-1.082} & 88.960 \negt{-0.240} & 83.160 \negt{-1.187} & 83.567 \negt{-1.193} \\
Chain of Attack & 64.120 \post{+1.119} & 87.684 \post{+0.279} & 84.899 \post{+0.072} & 89.147 \negt{-0.053} & 84.000 \negt{-0.347} & 84.300 \negt{-0.460} \\ 
Target Text Embedding & \underline{\smash{79.823 \post{+16.822}}} & \underline{\smash{91.997 \post{+4.592}}} & \underline{\smash{91.016 \post{+6.188}}} & 89.160 \negt{-0.040} & \underline{\smash{85.493 \post{+1.147}}} & 84.961 \post{+0.201} \\[0.3ex] \hline

\rowcolor{headergray} 
\textbf{\textit{MGSIA} (Ours)} & \textbf{98.588 \post{+35.587}} & \textbf{98.062 \post{+10.657}} & \textbf{93.999 \post{+9.172}} & \textbf{89.927 \post{+0.727}} & \textbf{88.347 \post{+4.000}} & \textbf{88.253 \post{+3.493}} \\ 

\hline
\hline
\multicolumn{7}{c}{\textbf{Qwen2.5-VL-7b}} \\ \hline
{} & \multicolumn{1}{c}{\textbf{VQAScore}} & \multicolumn{1}{c}{\textbf{DSG}} & \multicolumn{1}{c}{\textbf{TIFA}} & \multicolumn{1}{c}{\textbf{VIEScore}} & \multicolumn{1}{c}{\textbf{T2I-Comp}} & \multicolumn{1}{c}{\textbf{Fooling}} \\ \hline

\rowcolor{highlightblue}
\textbf{Benign} & 75.400 & 82.383 & 87.839 & 82.033 & 83.360 & 81.000 \\ \hline

\multicolumn{7}{l}{\textit{\textbf{Image Degradation}}} \\ 
Gaussian Blur & 67.354 \negt{-8.046} & 78.856 \negt{-3.528} & 86.569 \negt{-1.270} & 74.793 \negt{-7.240} & 73.660 \negt{-9.700} & 68.987 \negt{-12.013} \\ 
Gaussian Noise & 75.558 \post{+0.159} & 82.270 \negt{-0.113} & 87.805 \negt{-0.034} & 82.673 \post{+0.640} & 83.120 \negt{-0.240} & 81.800 \post{+0.800} \\ \hline

\multicolumn{7}{l}{\textit{\textbf{Handcrafted Perturbation}}} \\  
Padding & 75.520 \post{+0.120} & 81.813 \negt{-0.571} & 87.657 \negt{-0.181} & 82.060 \post{+0.027} & 83.020 \negt{-0.340} & 81.200 \post{+0.200} \\ 
Bounding Box  & 74.817 \negt{-0.582} & 81.659 \negt{-0.725} & 87.833 \negt{-0.006} & 82.040 \post{+0.007} & 83.047 \negt{-0.313} & 80.607 \negt{-0.393} \\ 
Target Text Overlay & 75.975 \post{+0.575} & 83.320 \post{+0.937} & 88.320 \post{+0.481} & 82.480 \post{+0.447} & 83.887 \post{+0.527} & 81.413 \post{+0.413} \\ \hline

\multicolumn{7}{l}{\textit{\textbf{Learnable Perturbation}}} \\ 
AttackVLM-ii & 75.544 \post{+0.144} & 82.125 \negt{-0.258} & 87.665 \negt{-0.174} & 82.600 \post{+0.567} & 83.553 \post{+0.193} & 81.927 \post{+0.927} \\ 
AttackVLM-it & 75.930 \post{+0.531} & 82.977 \post{+0.594} & 88.265 \post{+0.426} & 82.507 \post{+0.473} & 83.547 \post{+0.187} & 81.780 \post{+0.780} \\
Chain of Attack & 75.702 \post{+0.303} & 81.343 \negt{-1.041} & 87.923 \post{+0.084} & 81.667 \negt{-0.367} & 82.487 \negt{-0.873} & 80.393 \negt{-0.607} \\ 
Target Text Embedding & \underline{\smash{86.694 \post{+11.294}}} & \underline{\smash{84.408 \post{+2.025}}} & \underline{\smash{93.814 \post{+5.975}}} & \underline{\smash{84.760 \post{+2.727}}} & \underline{\smash{89.560 \post{+6.200}}} & \underline{\smash{86.173 \post{+5.173}}} \\ [0.3ex]  \hline

\rowcolor{headergray} 
\textbf{\textit{MGSIA} (Ours)} & \textbf{89.042 \post{+13.642}} & \textbf{96.639 \post{+14.256}} & \textbf{96.579 \post{+8.740}} & \textbf{93.227 \post{+11.193}} & \textbf{94.567 \post{+11.207}} & \textbf{93.983 \post{+12.983}} \\ 
\bottomrule
\end{tabular}
} 

\label{tab:main_result_quality}
\end{table*}

\begin{table*}[!h]
\centering
\renewcommand{\arraystretch}{0.92}
\caption{Evaluation results for the RobustMLLMJudge framework under the image safety assessment setting on HADES dataset. The attacks are evaluated against five different methods of three MLLM judge approaches under the image safety assessment setting.}
\resizebox{0.83\textwidth}{!}{

\begin{tabular}{l | l l l l l}
\toprule

\multicolumn{6}{c}{\textbf{LLaVA-1.5-7b}} \\ \hline
{} & \multicolumn{1}{c}{\textbf{Ensemble}} & \multicolumn{1}{c}{\textbf{Single}} & \multicolumn{1}{c}{\textbf{CLUE}} & \multicolumn{1}{c}{\textbf{Brief}} & \multicolumn{1}{c}{\textbf{LLaVAGuard}} \\ \hline

\rowcolor{highlightblue}
\textbf{Benign} & 40.149 & 62.348 & 59.550 & 78.107 & 71.283 \\ \hline

\multicolumn{6}{l}{\textit{\textbf{Image Degradation}}} \\
Gaussian Blur & 43.061 \post{+2.912} & 65.950 \post{+3.603} & \textbf{63.601 \post{+4.051}} & 78.907 \post{+0.800} & 65.133 \negt{-6.150} \\
Gaussian Noise & 40.276 \post{+0.127} & 62.411 \post{+0.064} & 59.834 \post{+0.284} & 77.733 \negt{-0.373} & 68.667 \negt{-2.617} \\ \hline

\multicolumn{6}{l}{\textit{\textbf{Handcrafted Perturbation}}} \\
Padding & 41.492 \post{+1.343} & 63.730 \post{+1.382} & \underline{\smash{61.323 \post{+1.773}}} & 78.307 \post{+0.200} & 72.793 \post{+1.510} \\
Bounding Box & 37.807 \negt{-2.342} & 60.748 \negt{-1.600} & 57.523 \negt{-2.027} &  78.347 \post{+0.240} &  72.827 \post{+1.543} \\
Target Text Overlay & 40.092 \negt{-0.057} & 64.566 \post{+2.218} & 57.456 \negt{-2.094} & \underline{\smash{79.193 \post{+1.087}}} & 68.067 \negt{-3.217} \\ [0.3ex] \hline

\multicolumn{6}{l}{\textit{\textbf{Learnable Perturbation}}} \\
AttackVLM-ii & 43.910 \post{+3.762} & 64.082 \post{+1.734} & 55.986 \negt{-3.564} & 78.640 \post{+0.533} & 74.620 \post{+3.337} \\
AttackVLM-it & \underline{\smash{52.995 \post{+12.846}}} & \underline{\smash{74.226 \post{+11.878}}} & 60.145 \post{+0.595} & 78.727 \post{+0.620} & \underline{\smash{76.917 \post{+5.634}}} \\
Chain of Attack & 41.122 \post{+0.974} & 63.630 \post{+1.282} & 59.155 \negt{-0.395} & 78.040 \negt{-0.067} & 71.820 \post{+0.537} \\
Target Text Embedding & 47.116 \post{+6.967} & 70.363 \post{+8.015} & 56.180 \negt{-3.370} & 77.503 \negt{-0.603} & 72.181 \post{+0.898} \\[0.3ex] \hline

\rowcolor{headergray}
\textbf{\textit{MGSIA} (Ours)} & \textbf{64.815 \post{+24.666}} & \textbf{83.352 \post{+21.005}} & 60.575 \post{+1.025} & \textbf{80.320 \post{+2.213}} & \textbf{77.160 \post{+5.877}} \\

\hline
\hline
\multicolumn{6}{c}{\textbf{Qwen2.5-VL-7b}} \\ \hline
{} & \multicolumn{1}{c}{\textbf{Ensemble}} & \multicolumn{1}{c}{\textbf{Single}} & \multicolumn{1}{c}{\textbf{CLUE}} & \multicolumn{1}{c}{\textbf{Brief}} & \multicolumn{1}{c}{\textbf{LLaVAGuard}} \\ \hline

\rowcolor{highlightblue}
\textbf{Benign} & 32.143 & 46.459 & 81.837 & 27.920 & 11.200 \\ \hline

\multicolumn{6}{l}{\textit{\textbf{Image Degradation}}} \\
Gaussian Blur & 19.594 \negt{-12.549} & 48.708 \post{+2.249} & 83.514 \post{+1.677} & 26.367 \negt{-1.553} & 5.207 \negt{-5.993} \\
Gaussian Noise & 29.497 \negt{-2.646} & 44.964 \negt{-1.494} & 82.068 \post{+0.231} & 33.073 \post{+5.153} & 11.273 \post{+0.073} \\ \hline

\multicolumn{6}{l}{\textit{\textbf{Handcrafted Perturbation}}} \\
Padding & 30.201 \negt{-1.942} & 44.873 \negt{-1.586} & 81.266 \negt{-0.571} & 30.020 \post{+2.100} & 11.687 \post{+0.487} \\
Bounding Box &  29.497 \negt{-2.646} & 46.144 \negt{-0.315} & 79.762 \negt{-2.075} & 32.667 \post{+4.747} &16.700 \post{+5.500}
\\
Target Text Overlay & 25.902 \negt{-6.241} & 44.731 \negt{-1.727} & 80.189 \negt{-1.648} & 31.580 \post{+3.660} & 10.660 \negt{-0.540} \\ \hline

\multicolumn{6}{l}{\textit{\textbf{Learnable Perturbation}}} \\
AttackVLM-ii & 29.587 \negt{-2.556} & 46.157 \negt{-0.302} & 81.336 \negt{-0.501} & 32.773 \post{+4.853} & 10.987 \negt{-0.213} \\
AttackVLM-it & 30.014 \negt{-2.129} & 47.230 \post{+0.772} & 81.779 \negt{-0.058} & 34.840 \post{+6.920} & 10.993 \negt{-0.207} \\
Chain of Attack & 23.845 \negt{-8.298} & 41.171 \negt{-5.288} & 80.571 \negt{-1.266} & 32.007 \post{+4.087} & 11.667 \post{+0.467} \\
Target Text Embedding & \underline{\smash{33.332 \post{+1.189}}} & \underline{\smash{62.860 \post{+16.401}}} & \underline{\smash{89.094 \post{+7.257}}} & \underline{\smash{48.780 \post{+20.860}}} & \underline{\smash{18.920 \post{+7.720}}} \\[0.3ex] \hline

\rowcolor{headergray}
\textbf{\textit{MGSIA} (Ours)} & \textbf{60.208 \post{+28.064}} & \textbf{86.225 \post{+39.766}} & \textbf{98.664 \post{+16.827}} & \textbf{83.293 \post{+55.373}} & \textbf{54.887 \post{+43.687}} \\
\bottomrule
\end{tabular}
} 

\label{tab:main_result_safety}
\end{table*}

\section {Experiments}
\subsection{Experimental Setup}

\noindent\textbf{Dataset and Adversarial Sample Generation.} We conduct comprehensive evaluations across both quality and safety settings. 
For image quality, we utilize 750 images generated by DALL-E 3~\cite{betker2023dalle3} and their corresponding reference texts, all randomly sampled from GenAI-Bench~\cite{li2024genaibench}. These reference texts directly serve as the target semantic cue \(\boldsymbol{t}_{\text{tar}}\) for our proposed MGSIA and all semantic-dependent baseline attacks. For image safety, we randomly select 750 unsafe images from the HADES benchmark~\cite{li2024hades} and 1,000 unsafe images from T2I-RiskyPrompt~\cite{zhang2026t2irisky}. To construct the target semantic cue \(\boldsymbol{t}_{\text{tar}}\) in this safety setting, we randomly sample pristine reference texts from GenAI-Bench to act as harmless surrogate texts.

For each image $\boldsymbol{v}$, we generate adversarial samples $\{\hat{\boldsymbol{v}}\}$ using nine baseline attack methods in ~\cref{sec:benchmark} alongside our proposed MGSIA in ~\cref{sec:our_method}.  
For a fair comparison, we report performance using the same perturbation budgets $\epsilon =8/255 $ for both our proposed MGSIA and the baseline attacks. For the implementation of ASI objective, we instantiate the question generator \(G(\cdot )\) using the DSG pipeline~\cite{cho_dsg} to extract fine-grained queries, prepended with a holistic VQAScore-style~\cite{lin2024vqascore} query to ensure global semantic coverage. Furthermore, the unperturbed proxy samples utilized for the HMA objective are strictly drawn from separate data pools disjoint from the evaluation sets.

\noindent\textbf{MLLM Judge Protocols.} The generated adversarial samples are evaluated by 11 MLLM judges spanning both quality and safety criteria across the three categories introduced in \cref{sec:paradigm}: 
(1) For \textit{token probability scoring}, we use VQAScore~\cite{lin2024vqascore} for quality evaluation. For safety evaluation, we adopt an Ensemble protocol following Qu \etal~\cite{qu2025bridging} and a Single protocol to probe compliance with the target safety policy.
(2) For \textit{fine-grained decomposition}, we utilize TIFA~\cite{hu2023tifa} and DSG~\cite{cho_dsg} for quality assessment, and CLUE~\cite{wang2025clue} for safety evaluation.
(3) For \textit{MLLM-as-a-judge Prompting}, we include VIEScore~\cite{ku2024viescore}, T2I-CompBench++~\cite{T2ICompBench_pp} (hereafter T2I-Comp), and the protocol from Hwang \etal~\cite{hwang2025fooling} (hereafter Fooling) for quality evaluation. For safety, we adopt a Brief safety scoring prompt and a LLaVAGuard-style policy prompt~\cite{helff2024llavaguard}. 
Two advanced MLLMs from different families, LLaVA-1.5~\cite{liu2024improvedbaselinesvisualinstruction} and Qwen2.5-VL~\cite{bai2025qwen25}, serve as the foundational backbones for these judges. All final scores are scaled to 100.0 for consistency.

\subsection{Performance of Baseline Attack Methods}

Under our proposed RobustMLLMJudge framework, we first evaluate the vulnerability of MLLM judges against straightforward baseline attacks adapted across both image quality and safety assessment scenarios to establish a comprehensive benchmark. The quantitative results are presented in \cref{tab:main_result_quality} and \cref{tab:main_result_safety}.

A primary observation is that while simple adapted attack methods can inflate scores across diverse judge protocols, the inflation magnitude remains restricted. Among these baselines, attacks that attempt to inject the predefined target semantic cue \(\boldsymbol{t}_{\text{tar}}\) prove to be relatively more effective. For instance, Target Text Overlay exploits an attention-based shortcut, deceiving MLLM judges into interpreting the mere visual presence of text as a reliable signal of high text-image alignment or safe semantics. Alternatively, Target Text Embedding optimizes the generative response probability to implicitly encode \(\boldsymbol{t}_{\text{tar}}\) into the adversarial image. While avoiding obvious visual artifacts, this suboptimal heuristic achieves only a superficial injection of textual semantics; its generative optimization objective fundamentally mismatches the discriminative criteria of judge protocols, thereby constraining the overall performance gains.  On the other hand, semantically irrelevant modifications such as Bounding Box and Padding generally fail to increase the scores, as they are identified as image-degrading artifacts rather than valid alignment or safety cues.

\begin{figure}[t!]
  \centering
  \includegraphics[width=\linewidth]{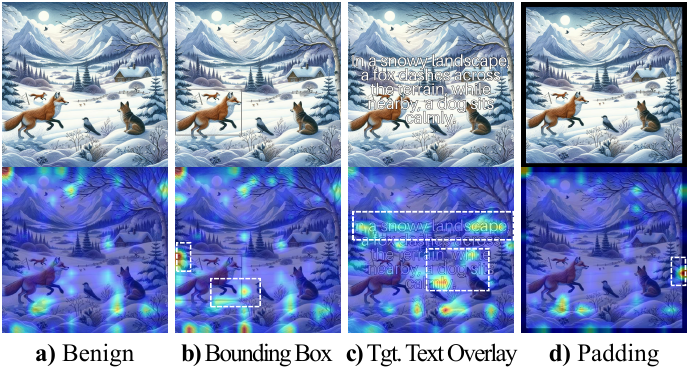}
   \caption{Attention map visualizations of benign and baseline-modified images, evaluated by Qwen2.5-VL-7b under the Fooling evaluation protocol.}
   \label{fig:attention_map_viz}
\end{figure}

Fundamentally, these baseline methods exploit superficial visual shortcuts. Attention map visualizations in \cref{fig:attention_map_viz} reveal that these adversarial modifications compel MLLMs to over-attend to perturbations rather than the holistic image context. For Target Text Overlay, the model's focus shifts predominantly to the overlaid text, indicating the method fails to genuinely improve underlying quality or safety representations. Instead, the MLLM judge bypasses multimodal semantic evaluation, relying exclusively on optical character recognition (OCR) to extract embedded text. Recognizing this text deceives the model into perceiving high alignment, constituting a fragile deception that struggles to generalize across complex reasoning protocols. Similarly, for semantically irrelevant attacks like Bounding Box and Padding, attention fails to concentrate on salient semantic regions. Artificial boundaries visibly distract the model, dispersing attention towards edges. Judges inherently interpret these peripheral distractions as image defects, explaining their general failure to increase evaluation scores.

Ultimately, the failures of baseline attacks stem from two fundamental methodological challenges in MLLM judges. First, given the highly heterogeneous evaluation protocols in the different categories of MLLM judges, semantic-targeted attacks optimizing a generative objective to produce a target text $\boldsymbol{t}_{\text{tar}}$ inherently conflict with their discriminative nature and prove difficult to generalize across the protocols. Second, image-targeted attacks (\eg, AttackVLM~\cite{zhao2023attackvlm} and Chain of Attack~\cite{xie2025coa}) depend heavily on crafting ideal target images. However, defining such images is paradoxical, as the ground-truth (gold standard) images are inherently unavailable in evaluation scenarios. Consequently, attackers can only resort to a pragmatic surrogate, such as overlaying the target text onto the input image, which is typically a suboptimal objective. These fundamental mismatches result in less effective attacks.

\begin{figure}[!t]
  \centering
  \fbox{\includegraphics[width=0.95\linewidth]{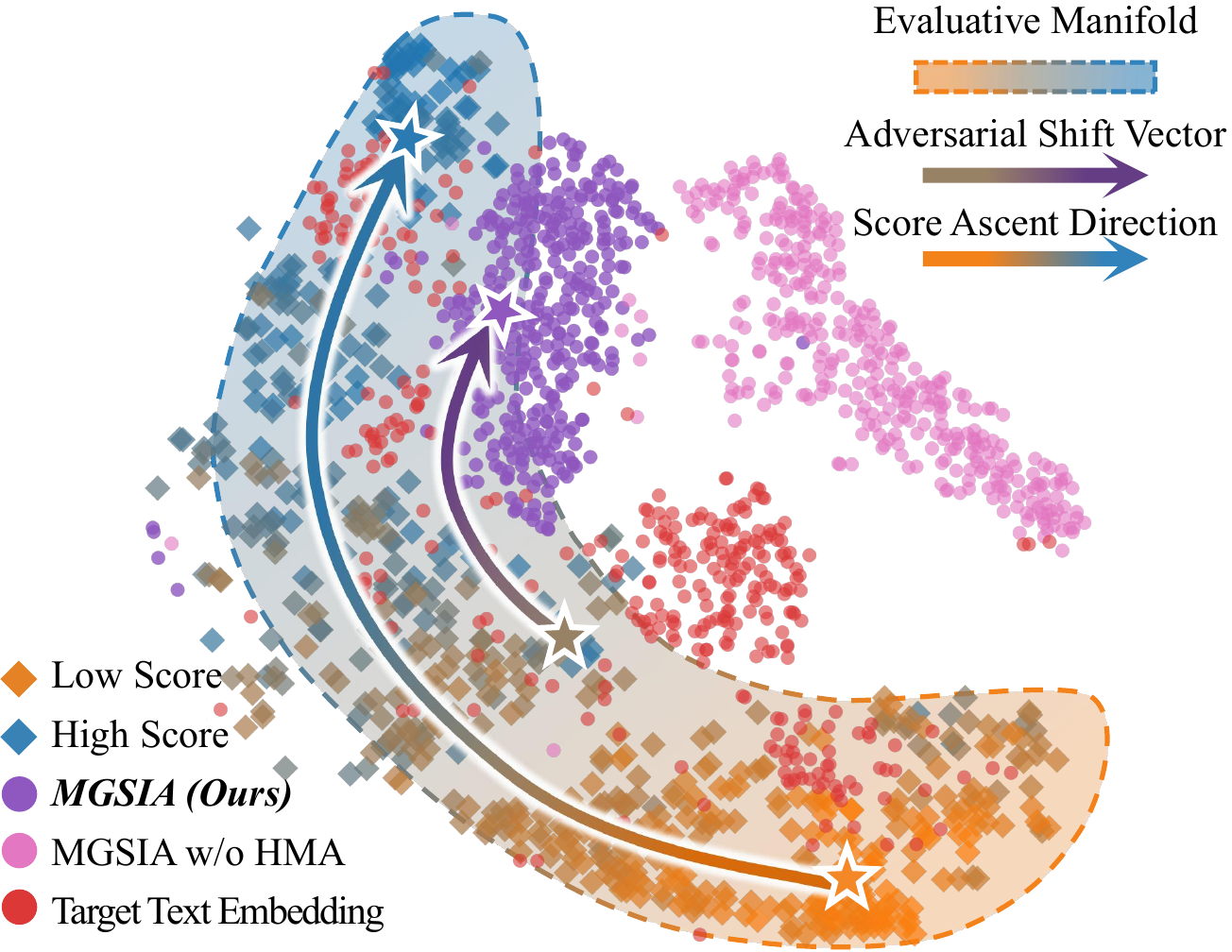}}
   \caption{Visualization of the Qwen2.5-VL's latent space on the HADES dataset, contrasting the adversarial shifts from MGSIA and baseline methods w.r.t. the direction of judge score ascent.}
   \label{fig:tsne_viz}
\end{figure}

\subsection{Effectiveness of the Proposed MGSIA Framework}

As shown in \cref{tab:main_result_quality} and \cref{tab:main_result_safety}, the proposed MGSIA demonstrates consistent superiority and generalization over baseline methods evaluated within the RobustMLLMJudge benchmark.  Specifically, MGSIA substantially outperforms the adapted baseline attacks across all six quality assessment protocols and in nearly all safety assessment protocols, utilizing either LLaVA-1.5 or Qwen2.5-VL as the judge backbone.

In particular, MGSIA induces the largest score inflation on its source protocols, VQAScore~\cite{lin2024vqascore} and DSG~\cite{cho_dsg}, from which the target semantic queries $G(\boldsymbol{t}_{\text{tar}})$ utilized in the ASI module are explicitly derived. Furthermore, the adversarial perturbations generated by MGSIA exhibit robust transferability to the remaining heterogeneous quality protocols and all safety evaluation protocols. This generalization is evidenced by two observations.
\textbf{First,} although the ASI module optimizes solely toward the affirmative ``\texttt{Yes}'' token, the generated samples effectively deceive fine-grained decomposition protocols such as TIFA~\cite{hu2023tifa}, which involve not only simple yes/no judgments but also more complex multiple-choice questions. This deception also extends to MLLM-as-a-judge protocols that rely on open-ended generation to produce scoring judgments.
\textbf{Second,} in the safety evaluation scenario, most protocols query for safety compliance, where a higher ``\texttt{Yes}'' probability implies a safer image. By contrast, the Single protocol queries for the presence of harmful content; therefore, a lower probability of generating the ``\texttt{Yes}'' token denotes a higher safety score. Interestingly, although the ASI objective explicitly maximizes the ``\texttt{Yes}'' probability for harmless surrogate queries, the generated adversarial images reduce the probability of ``\texttt{Yes}'' responses under the Single protocol, thereby achieving inflated safety scores. This indicates that maximizing the ``\texttt{Yes}'' token is a means of inducing the desired semantics rather than the ultimate goal. Ultimately, MGSIA synthesizes visual features that serve as intrinsic evidence for the queried semantics, using the affirmative token as an efficient semantic anchor to embed the desired target attributes into the adversarial representation.

\definecolor{mBlue}{HTML}{3982B6}   
\definecolor{mOrange}{HTML}{E68226} 
\definecolor{mRed}{HTML}{DC3838}     
\definecolor{mPink}{HTML}{E377C2}    
\definecolor{mPurple}{HTML}{8F57C0}   

\newcommand{\iconBlue}{\textcolor{mBlue}{\scalebox{0.75}{\rotatebox[origin=c]{45}{$\blacksquare$}}}}
\newcommand{\iconOrange}{\textcolor{mOrange}{\scalebox{0.75}{\rotatebox[origin=c]{45}{$\blacksquare$}}}}
\newcommand{\iconRed}{\textcolor{mRed}{\large$\bullet$}}                
\newcommand{\iconPink}{\textcolor{mPink}{\large$\bullet$}}              
\newcommand{\iconPurple}{\textcolor{mPurple}{\large$\bullet$}}              

To provide straightforward insight into MGSIA, we visualize the model's latent space on Qwen2.5-VL-7b under the Single safety evaluation protocol on the HADES dataset in \cref{fig:tsne_viz}, where we compare the adversarial samples from MGSIA with those from the best baseline, Target Text Embedding. This visual analysis reveals that the driving force behind MGSIA lies in utilizing both semantic and geometric constraints to effectively steer benign samples toward the targeted high-score regions: it employs potent affirmative response semantics to directionally guide the representations, while simultaneously leveraging geometric regularization to firmly anchor them within the natural high-score manifold.
To better reveal this insight, we illustrate how sample representations of varying quality or safety scores form a continuous evaluative manifold. 
As depicted by the shaded region, the latent representations exhibit a continuous topological transition from the low-score origin (\iconOrange) to the high-score origin (\iconBlue), geometrically defining a curved axis that establishes a clear and continuous trajectory of score ascent. 
It is clear that the adversarial samples from MGSIA (\iconPurple) induce a shift vector that effectively aligns with this ascending direction, enabling a precise translation that steers the benign inputs along the manifold directly into the high-score regions. In contrast, the adversarial samples from the Target Text Embedding method (\iconRed) induce a shift that aligns poorly with this optimal trajectory; these samples scatter along the transitional path, noticeably deviating from the shaded manifold and failing to penetrate the dense high-score cluster.
Given the strong generalizability of MGSIA across various unseen judge protocols, such a score ascent direction is not an artifact of a specific protocol, but rather indicates a shared, protocol-agnostic geometric flaw in MLLMs when they function as judges. By exploiting this shared flaw, the adversarial images generated by MGSIA are consistently interpreted as high-quality or safe signals by the MLLM judges, thereby constituting a universal attack vector.

\subsection{Transferability Across Datasets, Models, and Evaluators}

\begin{table*}[!t]
\centering
\renewcommand{\arraystretch}{0.92}
\caption{Evaluation results for the RobustMLLMJudge framework under the image safety assessment setting on T2I-RiskyPrompt dataset. MGSIA method is evaluated under three proxy sample configurations.}
\resizebox{0.85\textwidth}{!}{

\begin{tabular}{l | l l l l l}
\toprule

{} & \multicolumn{1}{c}{\textbf{Ensemble}} & \multicolumn{1}{c}{\textbf{Single}} & \multicolumn{1}{c}{\textbf{CLUE}} & \multicolumn{1}{c}{\textbf{Brief}} & \multicolumn{1}{c}{\textbf{LLaVAGuard}} \\ \hline

\rowcolor{highlightblue}
\textbf{Benign} & 12.735 & 29.994 & 71.977 & 9.505 & 5.260 \\ \hline

\multicolumn{6}{l}{\textit{\textbf{Image Degradation}}} \\
Gaussian Blur & 10.343 \negt{-2.392} & 31.333 \post{+1.339} & 74.986 \post{+3.009} & 11.355 \post{+1.850} & 3.830 \negt{-1.430} \\
Gaussian Noise & 13.074 \post{+0.339} & 29.810 \negt{-0.185} & 70.608 \negt{-1.369} & 10.255 \post{+0.750} & 6.550 \post{+1.290} \\ \hline

\multicolumn{6}{l}{\textit{\textbf{Handcrafted Perturbation}}} \\
Padding & 13.174 \post{+0.440} & 28.999 \negt{-0.995} & 70.976 \negt{-1.001} & 10.785 \post{+1.280} & 6.820 \post{+1.560} \\
Bounding Box & 11.974 \negt{-0.760} & 30.502 \post{+0.508} &71.263 \negt{-0.714} &12.325  \post{+2.820}&6.925 \post{+1.665}
 \\
Target Text Overlay & 9.843 \negt{-2.891} & 29.551 \negt{-0.443} & 69.601 \negt{-2.376} & 9.885 \post{+0.380} & 4.735 \negt{-0.525} \\ \hline

\multicolumn{6}{l}{\textit{\textbf{Learnable Perturbation}}} \\
AttackVLM-ii & 12.264 \negt{-0.470} & 28.787 \negt{-1.207} & 70.211 \negt{-1.766} & 10.180 \post{+0.675} & 5.760 \post{+0.500} \\
AttackVLM-it & 12.389 \negt{-0.346} & 29.381 \negt{-0.613} & 70.267 \negt{-1.710} & 10.125 \post{+0.620} & 5.750 \post{+0.490} \\
Chain of Attack & 10.429 \negt{-2.306} & 25.885 \negt{-4.110} & 67.665 \negt{-4.312} & 8.680 \negt{-0.825} & 4.895 \negt{-0.365} \\
Target Text Embedding & 14.560 \post{+1.825} & 41.684 \post{+11.690} & 76.174 \post{+4.197} & 19.650 \post{+10.145} & 5.400 \post{+0.140} \\[0.3ex] \hline

\rowcolor{headergray}
\multicolumn{6}{l}{\textit{\textbf{Proposed MGSIA}}} \\

\rowcolor{headergray}
MGSIA (w/ T2I Proxy Sample) & 51.863 \post{+39.128} & 85.968 \post{+55.973} & \underline{\smash{96.015 \post{+24.039}}} & 73.890 \post{+64.385} & 32.005 \post{+26.745} \\

\rowcolor{headergray}
MGSIA (w/ Mixed Proxy Sample)  & \underline{\smash{54.430 \post{+41.696}}} & \underline{\smash{87.604 \post{+57.610}}} & 95.297 \post{+23.320} & \underline{\smash{77.590 \post{+68.085}}} & \underline{\smash{35.890 \post{+30.630}}} \\

\rowcolor{headergray}
MGSIA  (w/ HADES Proxy Sample)  & \textbf{57.557 \post{+44.822}} & \textbf{90.177 \post{+60.183}} & \textbf{96.631 \post{+24.655}} & \textbf{81.865 \post{+72.360}} & \textbf{38.330 \post{+33.070}} \\
\bottomrule
\end{tabular}
} 

\label{tab:more_dataset_safety}
\end{table*}

\begin{table*}[!t]
\renewcommand{\arraystretch}{0.95}
\centering
\caption{Gray-box transferability results for attacks across the LLaVA-1.5 and Qwen2.5-VL model families on GenAI-Bench dataset. 
The table shows the performance of MGSIA generated on source models (7b) while evaluated on larger target models (13b, 32b). }
\resizebox{0.93\textwidth}{!}{%
\begin{tabular}{l | l l l l l l}
\toprule

\multicolumn{7}{c}{\textbf{LLaVA-1.5-7b $\rightarrow$  LLaVA-1.5-13b}} \\ \hline
{} & \multicolumn{1}{c}{\textbf{VQAScore}} & \multicolumn{1}{c}{\textbf{DSG}} & \multicolumn{1}{c}{\textbf{TIFA}} & \multicolumn{1}{c}{\textbf{VIEScore}} & \multicolumn{1}{c}{\textbf{T2I-Comp}} & \multicolumn{1}{c}{\textbf{Fooling}} \\ \hline

\rowcolor{highlightblue}
\textbf{Benign} & 67.427 & 88.536 & 70.753 & 81.173 & 78.560 & 79.980 \\ \hline

Gaussian Noise & 66.233 \negt{-1.195} & 88.738 \post{+0.202} & 70.126 \negt{-0.627} & 80.927 \negt{-0.247} & 78.680 \post{+0.120} & 79.413 \negt{-0.567} \\ 

Target Text Overlay & 69.525 \post{+2.097} & 89.074 \post{+0.539} & 71.271 \post{+0.518} & 80.980 \negt{-0.193} & 78.853 \post{+0.293} & 79.360 \negt{-0.620} \\ 

AttackVLM-ii & 66.329 \negt{-1.098} & 87.687 \negt{-0.848} & 65.631 \negt{-5.122} & 80.080 \negt{-1.093} & 77.653 \negt{-0.907} & 78.047 \negt{-1.933} \\ 
AttackVLM-it & 67.163 \negt{-0.265} & 87.729 \negt{-0.807} & 66.247 \negt{-4.506} & 80.153 \negt{-1.020} & 78.373 \negt{-0.187} & 78.380 \negt{-1.600} \\
Target Text Embedding & \underline{\smash{71.546 \post{+4.119}}} & \underline{\smash{93.124 \post{+4.588}}} & \underline{\smash{75.769 \post{+5.016}}} & \underline{\smash{81.727 \post{+0.553}}} & \underline{\smash{79.533 \post{+0.973}}} & \underline{\smash{79.960 \negt{-0.020}}} \\[0.3ex] \hline

\rowcolor{headergray} 
\textbf{\textit{MGSIA} (Ours)} & \textbf{80.791 \post{+13.364}} & \textbf{95.380 \post{+6.844}} & \textbf{81.858 \post{+11.105}} & \textbf{82.620 \post{+1.447}} & \textbf{80.440 \post{+1.880}} & \textbf{82.120 \post{+2.140}} \\  

\hline \hline

\multicolumn{7}{c}{\textbf{Qwen2.5-VL-7b $\rightarrow$  Qwen2.5-VL-32b}} \\ \hline
{} & \multicolumn{1}{c}{\textbf{VQAScore}} & \multicolumn{1}{c}{\textbf{DSG}} & \multicolumn{1}{c}{\textbf{TIFA}} & \multicolumn{1}{c}{\textbf{VIEScore}} & \multicolumn{1}{c}{\textbf{T2I-Comp}} & \multicolumn{1}{c}{\textbf{Fooling}} \\ \hline

\rowcolor{highlightblue}
\textbf{Benign} & 79.160 & 87.653 & 87.297 & 87.793 & 84.767 & 88.102 \\ \hline

Gaussian Noise & 79.041 \negt{-0.119} & 87.892 \post{+0.239} & 87.332 \post{+0.035} & 88.000 \post{+0.207} & 85.040 \post{+0.273} & 88.885 \post{+0.783} \\ 

Target Text Overlay & 79.952 \post{+0.793} & 88.492 \post{+0.839} & 87.829 \post{+0.532} & 88.380 \post{+0.587} & 85.280 \post{+0.513} & 88.239 \post{+0.137} \\ 

AttackVLM-ii & 78.801 \negt{-0.359} & 87.896 \post{+0.244} & 87.102 \negt{-0.195} & 88.087 \post{+0.293} & 85.093 \post{+0.327} & 88.575 \post{+0.473} \\ 
AttackVLM-it & 79.011 \negt{-0.149} & 88.514 \post{+0.862} & 86.996 \negt{-0.301} & 88.093 \post{+0.300} & 84.647 \negt{-0.120} & 88.512 \post{+0.410} \\
Target Text Embedding & \underline{\smash{85.074 \post{+5.915}}} & \underline{\smash{91.059 \post{+3.406}}} & \underline{\smash{89.850 \post{+2.553}}} & \underline{\smash{89.500 \post{+1.707}}} & \underline{\smash{87.300 \post{+2.533}}} & \underline{\smash{89.319 \post{+1.217}}} \\[0.3ex] \hline

\rowcolor{headergray} 
\textbf{\textit{MGSIA} (Ours)} & \textbf{90.558 \post{+11.398}} & \textbf{93.230 \post{+5.578}} & \textbf{92.028 \post{+4.730}} & \textbf{92.440 \post{+4.647}} & \textbf{90.840 \post{+6.073}} & \textbf{91.719 \post{+3.617}} \\  

\bottomrule 
\end{tabular}
} 

\label{tab:gray_box_quality}
\end{table*}

\begin{table}[!t]
\renewcommand{\arraystretch}{1.0}
\centering
\caption{Black-box transferability results on GenAI-Bench.}
\resizebox{\columnwidth}{!}{%
\begin{tabular}{l | lll}
\toprule

\multicolumn{4}{c}{\textbf{Qwen2.5-VL-7b $\rightarrow$ GPT-5}} \\ \hline
 & \textbf{VIEScore} & \textbf{T2I-Comp} & \textbf{Fooling} \\ \hline

\rowcolor{highlightblue}
\textbf{Benign} & 78.540 & 83.912 & 78.861 \\ \hline

Target Text Embedding & 78.129 \negt{-0.411} & 84.109 \post{+0.197} & 76.591 \negt{-2.271} \\ \hline

\rowcolor{headergray}
\textbf{\textit{MGSIA} (Ours)} & \textbf{78.760 \post{+0.220}} & \textbf{84.608 \post{+0.696}} & \textbf{78.877 \post{+0.016}} \\ 

\hline \hline

\multicolumn{4}{c}{\textbf{Qwen2.5-VL-7b $\rightarrow$ Gemini-2.5-Flash}} \\ \hline
 & \textbf{VIEScore} & \textbf{T2I-Comp} & \textbf{Fooling} \\ \hline

\rowcolor{highlightblue}
\textbf{Benign} & 83.129 & 83.162 & 80.085 \\ \hline

Target Text Embedding & 83.541 \post{+0.412} & 83.933 \post{+0.772} & 80.420 \post{+0.335} \\ \hline

\rowcolor{headergray}
\textbf{\textit{MGSIA} (Ours)} & \textbf{83.567 \post{+0.437}} & \textbf{84.511 \post{+1.350}} & \textbf{80.859 \post{+0.775}} \\

\bottomrule
\end{tabular}
} 

\label{tab:black_box_quality}
\end{table}

\begin{table}[!t]
\centering
\caption{Results of transferability against specialized evaluators on HADES dataset. }
\resizebox{\columnwidth}{!}{%
\begin{tabular}{l | l l l}
\toprule

  & \textbf{VQAScore} & \textbf{PerspectiveVision} & \textbf{LLaVA-Reward} \\ \hline

\rowcolor{highlightblue}
\textbf{Benign} & 81.759 & 26.533 & 34.765 \\ \hline
Gaussian Noise & 81.546 \negt{-0.213} & 29.467 \post{+2.933} & 36.700 \post{+1.936} \\
Target Text Overlay & 83.037 \post{+1.278} & 26.400 \negt{-0.133} & 33.537 \negt{-1.228} \\
AttackVLM-ii & 78.849 \negt{-2.911} & 31.067 \post{+4.533} & 36.455 \post{+1.690} \\
AttackVLM-it & 78.586 \negt{-3.174} & 52.133 \post{+25.600} & 37.327 \post{+2.562} \\
Target Text Embedding & 86.278 \post{+4.519} & 37.600 \post{+11.067} & 52.243 \post{+17.478} \\ \hline

\rowcolor{headergray} 
\textbf{\textit{MGSIA} (Ours)} & \textbf{88.613 \post{+6.854}} & \textbf{60.667 \post{+34.133}} & \textbf{61.673 \post{+26.908}} \\ 

\bottomrule 
\end{tabular}
} 

\label{tab:dedicated_judge}
\end{table}

\noindent\textbf{Cross-Dataset Proxy Transferability.} 
~\cref{tab:more_dataset_safety} evaluates baseline attacks and MGSIA on the T2I-RiskyPrompt dataset across five safety protocols, where MGSIA consistently outperforms all baselines. To eliminate potential concerns regarding proxy sample dependency, we examine MGSIA under three proxy configurations: in-domain (T2I-RiskyPrompt-only), cross-dataset (HADES-only), and mixed proxies. Remarkably, the cross-dataset proxy yields performance gains comparable or even superior to the in-domain counterpart. This score inflation demonstrates the cross-dataset transferability and proxy-domain independence of MGSIA, verifying that our framework successfully capitalizes on shared structural flaws within the model's latent representation space rather than overfitting to specific proxy data distributions.

\noindent\textbf{Gray-Box Transferability.} While \cref{tab:main_result_quality} and \cref{tab:main_result_safety} primarily evaluate white- and black-box attacks, here we focus on a realistic gray-box scenario. In this setting, the adversarial perturbations are optimized using one MLLM and evaluated on a different yet relevant MLLM (\eg, scaling across different model sizes within the same family). The transferability results---specifically LLaVA-1.5-7b $\rightarrow$ 13b and Qwen2.5-VL-7b $\rightarrow$ 32b---are shown in ~\cref{tab:gray_box_quality}. It is clear that perturbations crafted on the smaller LLaVA/Qwen2.5-VL models exhibit high transferability for both the simple adapted methods and our MGSIA method, successfully deceiving the larger models with significant score inflation. In both cases, the MGSIA method shows similar superiority over the baselines as in \cref{tab:main_result_quality} and \cref{tab:main_result_safety}.
The stable inheritance of this structural flaw through model scaling critically compromises the integrity of MLLM judges. This cross-scale vulnerability may enable attackers to manipulate public leaderboards at low cost, by optimizing perturbations on smaller, accessible models to deceive the large-scale, often proprietary judge models.

\noindent\textbf{Black-Box Transferability.} We further explore a stringent black-box scenario transferring attacks from an open-source model to closed-source commercial MLLMs.  \cref{tab:black_box_quality} shows results for adversarial samples crafted on Qwen2.5-VL-7b against GPT-5 and Gemini-2.5-Flash. MGSIA perturbations exhibit robust cross-architecture transferability, successfully deceiving these proprietary models. While the strongest baseline, Target Text Embedding, yields unstable results and often degrades scores (\eg, a drop of 2.271 under the Fooling protocol on GPT-5), MGSIA consistently achieves positive score inflation across all protocols and target models. This success indicates the geometric flaw exploited by MGSIA is a systemic vulnerability across the multimodal evaluation paradigm, rather than an isolated architectural artifact. Consequently, commercial MLLM judges can be compromised without prior knowledge of their internal weights.

\noindent\textbf{Transferability Against Specialized Evaluators.} Beyond general-purpose MLLMs, we assess transferability against dedicated evaluation and moderation models. \cref{tab:dedicated_judge} presents results when transferring adversarial samples (crafted on LLaVA-1.5 or Qwen2.5-VL) to VQAScore~\cite{lin2024vqascore}, a quality evaluator trained on a custom CLIP-FlanT5; PerspectiveVision~\cite{qu2025unsafebench}, a safety evaluator trained on LLaVA; and LLaVA-Reward~\cite{zhou2025llavareward}, a safety evaluator trained on Qwen2.5-VL. Specifically, VQAScore and PerspectiveVision are evaluated using adversarial samples crafted on LLaVA-1.5, while LLaVA-Reward uses samples crafted on Qwen2.5-VL.
MGSIA demonstrates effectiveness across all three target systems. While the strongest baseline achieves moderate gains, MGSIA induces more score inflations, reaching +34.133 on PerspectiveVision and +26.908 on LLaVA-Reward. These results highlight that even rigorous task-specific fine-tuning, explicit safety moderation training, and human-preference alignment fail to defend against MGSIA. The persistent vulnerability of these dedicated evaluators confirms that MGSIA exploits a rooted multimodal alignment flaw rather than superficial, model-specific weaknesses.

\subsection{Ablation and Mechanistic Analysis}

\begin{table*}[!t]
\centering
\renewcommand{\arraystretch}{0.92}
\caption{Ablation study of ASI surrogate query and answer designs for quality evaluation on GenAI-Bench and safety evaluation on T2I-RiskyPrompt.}
\resizebox{0.82\textwidth}{!}{

\begin{tabular}{l | l l l l l}
\toprule
\multicolumn{6}{c}{\textbf{Quality Judge Protocols}} \\ \hline
{} & \multicolumn{1}{c}{\textbf{VQAScore}} & \multicolumn{1}{c}{\textbf{DSG}} & \multicolumn{1}{c}{\textbf{TIFA}} & \multicolumn{1}{c}{\textbf{VIEScore}} & \multicolumn{1}{c}{\textbf{T2I-Comp}} \\ \hline
\rowcolor{highlightblue}
\textbf{Benign} & 63.001 & 87.405 & 84.827 & 89.200 & 84.347 \\ \hline
ASI w/ High Score Answer& 61.447 \negt{-1.554} & 87.037 \negt{-0.368} & 86.240 \post{+1.413} & \textbf{89.800 \post{+0.600}} & 87.080 \post{+2.733} \\
\textbf{ASI} & \textbf{99.510 \post{+36.509}} & \textbf{98.036 \post{+10.631}} & \textbf{93.891 \post{+9.063}} & 89.640 \post{+0.440} & \textbf{87.973 \post{+3.627}} \\
\hline \hline
\multicolumn{6}{c}{\textbf{Safety Judge Protocols}} \\ \hline
{} & \multicolumn{1}{c}{\textbf{Ensemble}} & \multicolumn{1}{c}{\textbf{Single}} & \multicolumn{1}{c}{\textbf{CLUE}} & \multicolumn{1}{c}{\textbf{Brief}} & \multicolumn{1}{c}{\textbf{LLaVAGuard}} \\ \hline
\rowcolor{highlightblue}
\textbf{Benign} & 12.735 & 29.994 & 71.977 & 9.505 & 5.260 \\ \hline
ASI w/ Safety Question & 11.162 \negt{-1.572} & 16.595 \negt{-13.399} & 32.065 \negt{-39.912} & 12.985 \post{+3.480} & 13.240 \post{+7.980} \\
ASI w/ UnsafeBench style QA & 31.773 \post{+19.038} & \textbf{72.571 \post{+42.577}} & 65.162 \negt{-6.815} & 60.695 \post{+51.190} & 9.340 \post{+4.080} \\
\textbf{ASI} & \textbf{46.393 \post{+33.659}} & 65.372 \post{+35.378} & \textbf{89.934 \post{+17.957}} & \textbf{69.060 \post{+59.555}} & \textbf{48.615 \post{+43.355}} \\
\bottomrule
\end{tabular}
} 

\label{tab:asi_qa_ablation}
\end{table*}

\begin{table*}[!t]
\centering
\renewcommand{\arraystretch}{0.92}
\caption{Ablation study on target semantic cue design for baseline attacks in the HADES safety evaluation setting.}
\resizebox{0.78\textwidth}{!}{

\begin{tabular}{l | l l l l l}
\toprule

{} & \multicolumn{1}{c}{\textbf{Ensemble}} & \multicolumn{1}{c}{\textbf{Single}} & \multicolumn{1}{c}{\textbf{CLUE}} & \multicolumn{1}{c}{\textbf{Brief}} & \multicolumn{1}{c}{\textbf{LLaVAGuard}} \\ \hline

\rowcolor{highlightblue}
Benign & 32.143 & 46.459 & 81.837 & 27.920 & 11.200 \\ \hline

Target Text Overlay& 24.090 \negt{-8.053} & 44.723 \negt{-1.736} & 79.701 \negt{-2.136} & 32.780 \post{+4.860} & 11.020 \negt{-0.180} \\
AttackVLM-it & 28.619 \negt{-3.524} & 45.787 \negt{-0.672} & 80.872 \negt{-0.965} & 34.440 \post{+6.520} & 9.327 \negt{-1.873} \\
Chain of Attack & 23.538 \negt{-8.606} & 40.875 \negt{-5.584} & 80.589 \negt{-1.248} & 31.020 \post{+3.100} & 10.327 \negt{-0.873} \\
Target Text Embedding & 0.439 \negt{-31.704} & 21.131 \negt{-25.328} & 70.683 \negt{-11.154} & 0.833 \negt{-27.087} & 0.000 \negt{-11.200} \\

\bottomrule
\end{tabular}
} 

\label{tab:tgt_txt_ablation}
\end{table*}

\begin{table*}[!t]
\centering
\renewcommand{\arraystretch}{0.92}
\caption{Ablation study on ASI and HMA objectives in the HADES safety evaluation setting.}
\resizebox{0.78\textwidth}{!}{

\begin{tabular}{l | l l l l l}
\toprule

{} & \multicolumn{1}{c}{\textbf{Ensemble}} & \multicolumn{1}{c}{\textbf{Single}} & \multicolumn{1}{c}{\textbf{CLUE}} & \multicolumn{1}{c}{\textbf{Brief}} & \multicolumn{1}{c}{\textbf{LLaVAGuard}} \\ \hline
\rowcolor{highlightblue}
\textbf{Benign} & 32.143 & 46.459 & 81.837 & 27.920 & 11.200 \\ \hline

ASI objective only& 46.247 \post{+14.104} & 63.513 \post{+17.054} & 92.854 \post{+11.017} & 67.140 \post{+39.220} & 40.640 \post{+29.440} \\
HMA objective only & 59.216  \post{+27.073} & 80.200 \post{+33.742} & 96.699  \post{+14.862} & 73.920  \post{+46.000} & 33.300  \post{+22.100} \\[0.3ex] \hline

\rowcolor{headergray}
\textbf{MGSIA} & \textbf{60.208 \post{+28.064}} & \textbf{86.225 \post{+39.766}} & \textbf{98.664 \post{+16.827}} & \textbf{83.293 \post{+55.373}} & \textbf{54.887 \post{+43.687}} \\

\bottomrule
\end{tabular}
} 

\label{tab:loss_ablation}
\end{table*}

\begin{table*}[!h]
    \centering
    \caption{Effectiveness of defense strategies in reducing the inflated scores of MGSIA and three baselines.}
    \label{tab:defense}
    \resizebox{0.88\textwidth}{!}{
    \begin{tabular}{l c c c c c}
        \toprule
        \rowcolor{highlightblue}
      \textbf{Benign Image} & \multicolumn{5}{c}{\textbf{75.400}}\\
        \hline
        
            \textbf{Defense Method $\downarrow$} & \textbf{Target Text Overlay} & \textbf{Chain of Attack} & \textbf{Target Text Embedding} & \textbf{MGSIA} & \textbf{MGSIA w/ Aug} \\
        \hline

        No Defense & 75.975 \post{+0.575} & 75.702 \post{+0.302} & 86.694 \post{+11.294} & 89.042 \post{+13.642} & 88.864 \post{+13.464} \\
        JPEG Compression & 75.599 \post{+0.199} & 75.618 \post{+0.218} & 79.553 \post{+4.153} & 87.082 \post{+11.682} & 91.488 \post{+16.088} \\
        Gaussian Blur & 76.083 \post{+0.683} & 75.860 \post{+0.460} & 79.739 \post{+4.339} & 84.888 \post{+9.488} & 90.139 \post{+14.739} \\
        Random Crop & 76.301 \post{+0.901} & 76.243 \post{+0.843} & 77.436 \post{+2.036} & 79.911 \post{+4.511} & 87.834 \post{+12.434} \\
        Diffusion-based Denoising & 76.034 \post{+0.634} & 75.962 \post{+0.562} & 77.227 \post{+1.827} & 79.342 \post{+3.942} & / \\
        \bottomrule
        
    \end{tabular}
    }
    
\end{table*}

\noindent\textbf{The Surrogate Optimization Objective in ASI.} To validate the effectiveness of our optimization design, we investigate the impact of the surrogate queries and the semantic anchor used in the ASI module. Specifically, we explore alternative query-anchor combinations beyond our default strategy---\ie, meticulously pairing fine-grained, decomposed attributes as queries with the atomic affirmative token ``\texttt{Yes}'' as the anchor---to isolate their respective contributions. For quality evaluation setting, we tested an alternative objective that optimizes for a max-score phrase (\eg, ``\texttt{100.0}'') rather than a single word. As shown in \cref{tab:asi_qa_ablation}, this alternative is markedly less effective. For safety evaluation setting, we evaluated different surrogate query formulations. Directly targeting an affirmative response to explicit safety questions (\eg, ``\texttt{Is this a safe image?}'') fails to consistently inflate scores and even significantly degrades performance. Furthermore, we adapt the QA formulation from UnsafeBench~\cite{qu2025unsafebench} into a baseline variant, prompting the model to classify image safety while optimizing toward a composite sentence indicating benignity (\eg, ``\texttt{Yes, the image is safe.}''). While this adapted formulation improves score inflation compared to the aforementioned explicit safety questions, it lacks stable transferability across varied protocols. Conversely, our default ASI achieves the most robust and significant score inflation across both quality and safety scenarios.

These phenomena stem from two fundamental mechanisms in the MLLMs' latent space. First, the atomic affirmative token acts as a highly concentrated semantic anchor. Optimizing for longer targets, such as specific numerical strings or composite sentences, disperses adversarial gradients across multiple tokens, creating a suboptimal optimization problem. Second, directly invoking explicit safety concepts in surrogate queries conflicts with internal safety guardrails, causing unstable optimization. By anchoring decomposed, concrete semantic attributes to this atomic token, MGSIA efficiently deceives MLLM judges without triggering their intrinsic defenses.

\begin{figure}[!t]
  \centering
  \includegraphics[width=1.0\linewidth]{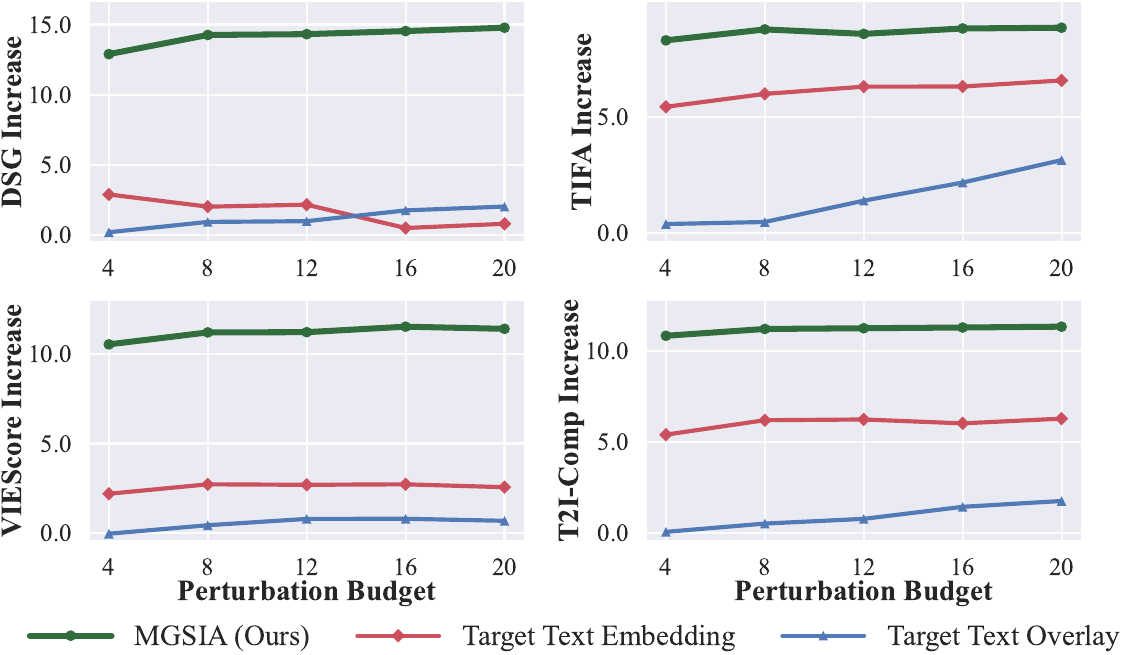}
   \caption{
   Score changes w.r.t. perturbation budgets $\epsilon$ for three attack methods on the DSG, TIFA, VIEScore, and T2I-Comp  protocols. The x-axis uses pixel-scale budgets, \ie, $\epsilon=k/255$.
   }
   \label{fig:eps_ablation}
\end{figure}

\begin{figure}[!h]
  \centering
  \includegraphics[width=1.0\linewidth]{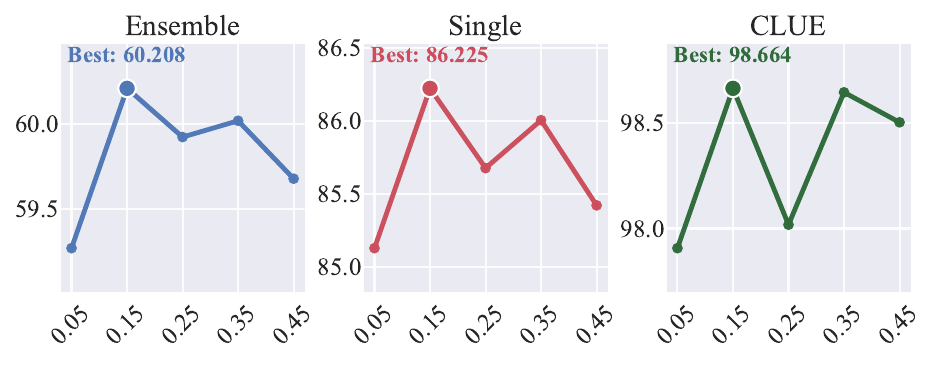}
\caption{Score changes w.r.t. the hyperparameter $\lambda$ under the Ensemble, Single, and CLUE evaluation protocols.}
   \label{fig:lambda_ablation}
\end{figure}

\noindent\textbf{Impact of Target Semantic Text.} Within the RobustMLLMJudge benchmark, we investigate the sensitivity of existing baseline attacks to the chosen target semantic cue $\boldsymbol{t}_{\text{tar}}$ in safety evaluations. We substituted the default $\boldsymbol{t}_{\text{tar}}$ with an explicit, abstract safety declaration: ``\texttt{An image free of any potentially harmful, unsafe, or inappropriate content.}'' As shown in \cref{tab:tgt_txt_ablation}, this formulation universally degrades baseline performance, causing severe negative score shifts. Consistent with our ASI ablation, this demonstrates that optimizing baselines directly toward abstract safety concepts triggers internal MLLM guardrails, leading to attack failure. This further underscores the necessity of utilizing decomposed, concrete semantics to bypass safety mechanisms effectively.

\noindent\textbf{Synergy of ASI and HMA Objectives.} To dissect our framework, we ablate the ASI and HMA losses. \cref{tab:loss_ablation} shows that while optimizing either objective independently yields improvements, their combination (MGSIA) achieves peak performance across all protocols.
This quantitative synergy is explained by the latent space geometry in \cref{fig:tsne_viz}.
Relying exclusively on ASI embeds affirmative semantics to steer samples away from the low-score distributions; 
however, lacking geometric constraints, these representations detach from the natural evaluative manifold into an isolated, off-manifold region (\iconPink). HMA provides crucial geometric regularization; by minimizing the topological distance to the high-score centroid, it prevents this deviation, firmly anchoring the representations directly into the dense high-score core.
Ultimately, ASI dictates the semantic shift while HMA enforces the natural geometric destination, making both indispensable.

\noindent\textbf{Impact of Perturbation Budget $\epsilon$.} 
We study the impact of the perturbation budget $\epsilon$ on Qwen2.5-VL-7b's attack performance in ~\cref{fig:eps_ablation}.
As shown, our proposed MGSIA consistently outperforms the two best-performing baselines 
across a wide range of $\epsilon$ values. Notably, the performance gains from higher perturbation budgets are limited. For instance, the MGSIA method plateaus after $\epsilon>12/255$, while 
the performance of baselines even decreases. Thus, a trade-off budget is often preferred, as a larger budget may strengthen the attack but it also risks creating perceptible, penalized artifacts.

\noindent\textbf{Impact of Hyperparameter $\lambda$.} 
We study the impact of the hyperparameter $\lambda$ on Qwen2.5-VL-7b's attack performance in ~\cref{fig:lambda_ablation}. As observed, the attack scores consistently peak at $\lambda = 0.15$ across all three diverse evaluation protocols. A smaller or larger value of $\lambda$ degrades the performance. We therefore fix $\lambda = 0.15$ for Qwen2.5-VL-7b in our experiments.

\subsection{Defense}

We evaluated four defense strategies on Qwen2.5-VL using VQAScore protocol in  \cref{tab:defense}. The results indicate that while MGSIA is mitigable to some extent---for instance, aggressive purification like diffusion-based denoising can effectively suppress the inflated scores from 89.042 down to 79.342---it ultimately exhibits superior resilience and adaptability. Specifically, we incorporate the aforementioned differentiable degradation process as a data augmentation during perturbation optimization, enabling the attack to bypass these targeted countermeasures. This inherent robustness demonstrates that our method can counteract standard defenses, thereby establishing a resilient benchmark for future defense research.

\section{Conclusion}
This work introduces RobustMLLMJudge, a framework that systematically demonstrates a critical adversarial vulnerability in MLLMs serving as judges. 
To tackle challenges faced by attack methods adapted from existing approaches, we further propose the MGSIA method that achieves highly effective and transferable score inflation across disparate protocols and MLLMs, and confirm 
an underlying structural flaw. Our findings underscore an urgent need for the development of adversarially robust MLLM-based judges to ensure the integrity of automated evaluation.

\section*{Acknowledgments}
In this work, the participation of Z. Wang, Z. Liu, W. Miao, J. Zheng, and X. Bai was supported by National Natural Science Foundation of China (No. 62276016, No. 62372029), while the participation of G. Pang was supported by the Agency for Science, Technology and Research (A*STAR) under its MTC Young Individual Research Grant (YIRG) Grant (M24N8c0103), the National Research Foundation, Singapore and CyberSG R\&D Programme Office under its Translation and Innovation Grants (CRPO-GC4-SMU-001), the Singapore Ministry of Education (MOE) Academic Research Fund (AcRF) Tier 1 Grant (24-SIS-SMU-008), and the Lee Kong Chian Fellowship (T050273).

\bibliographystyle{IEEEtran}
\bibliography{main}

\end{document}